\documentclass[12pt]{article}
\usepackage{amsfonts,amsmath,amssymb,amsthm}
\usepackage{graphicx,psfrag,epsf}
\usepackage{enumerate}
\usepackage{enumitem}
\usepackage{url}
\usepackage{algorithm}
\usepackage{algpseudocode}
\usepackage{authblk}
\usepackage{helvet}
\usepackage[colorlinks=true,pagebackref,linkcolor=magenta]{hyperref}
\usepackage[sort&compress,comma,square,numbers]{natbib}

\usepackage{color} 
\usepackage{paralist}
\usepackage{lineno}
\usepackage[font=small,labelfont=bf]{caption}
\usepackage{chngcntr}

\usepackage{tikz}
\usepackage{graphicx}
\usetikzlibrary{calc,shapes.geometric, arrows,positioning}
\tikzstyle{node} = [rectangle, rounded corners, minimum width=1cm, minimum height=1cm]
\tikzstyle{arrow} = [thick,->,>=stealth]

\providecommand{\sct}[1]{{\normalfont\textsc{#1}}}

\providecommand{\mb}[1]{\boldsymbol{#1}}
\providecommand{\mc}[1]{\mathcal{#1}}
\newcommand{\Real}{\mathbb{R}}
\newcommand{\GG}{c}

\newcommand{\E}{\hat{E}}

\newcommand{\TT}{^{\ensuremath{\mathsf{T}}}}

\newcommand{\Mgc}{\sct{MGC}}

\newcommand{\Hhg}{\sct{Hhg}}
\newcommand{\Hsic}{\sct{Hsic}}
\newcommand{\Dcorr}{\sct{Dcorr}}

\newcommand{\Mantel}{\sct{Mantel}}
\newcommand{\Pearson}{\sct{Pearson}}
\newcommand{\Copula}{\sct{Copula}}

\newcommand{\mbx}{\ensuremath{X}}
\newcommand{\mby}{\ensuremath{Y}}

\newtheorem{thm}{Theorem}
\newtheorem*{thm*}{Theorem}
\newtheorem*{defi*}{Definition}
\newtheorem{cor}{Corollary}
\newtheorem{lem}{Lemma}

\newcommand{\blind}{1}

\begin{document}

\def\spacingset#1{\renewcommand{\baselinestretch}%
{#1}\small\normalsize} \spacingset{1}

\if1\blind
{
  \title{\bf From Distance Correlation to Multiscale Graph Correlation}
  \author[1]{Cencheng Shen\thanks{shenc@udel.edu}}
\author[2]{Carey E. Priebe\thanks{cep@jhu.edu}}
\author[3]{Joshua T. Vogelstein\thanks{jovo@jhu.edu}}
\affil[1]{Department of Applied Economics and Statistics, University of Delaware}
\affil[2]{Department of Applied Mathematics and Statistics, Johns Hopkins University}
\affil[3]{Department of Biomedical Engineering and Institute of Computational Medicine, Johns Hopkins University}
  \maketitle
} \fi

\if0\blind
{
  \bigskip
  \bigskip
  \bigskip
  \begin{center}
    {\LARGE\bf From Distance Correlation to Multiscale Graph Correlation}
\end{center}
  \medskip
} \fi

\bigskip
\begin{abstract}
Understanding and developing a correlation measure that can detect general dependencies is not only imperative to statistics and machine learning, but also crucial to general scientific discovery in the big data age. In this paper, we establish a new framework that generalizes distance correlation --- a correlation measure that was recently proposed and shown to be universally consistent for dependence testing against all joint distributions of finite moments --- to the Multiscale Graph Correlation (\Mgc). By utilizing the characteristic functions and incorporating the nearest neighbor machinery, we formalize the population version of local distance correlations, define the optimal scale in a given dependency, and name the optimal local correlation as \Mgc. The new theoretical framework motivates a theoretically sound Sample \Mgc~and allows a number of desirable properties to be proved, including the universal consistency, convergence and almost unbiasedness of the sample version. The advantages of \Mgc~are illustrated via a comprehensive set of simulations with linear, nonlinear, univariate, multivariate, and noisy dependencies, where it loses almost no power in monotone dependencies while achieving better performance in general dependencies, compared to distance correlation and other popular methods.
\end{abstract}

\noindent%
\vfill



\newpage
\spacingset{1.45} 
 
\section{Introduction}
Given pairs of observations $(x_{i},y_{i}) \in \Real^{p} \times \Real^{q}$ for $i=1,\ldots,n$, assume they are generated by independently identically distributed (\emph{iid}) $F_{\mbx \mby}$. 
A fundamental statistical question prior to the pursuit of any meaningful joint inference is the independence testing problem: the two random variables are independent if and only if $F_{\mbx \mby} = F_{\mbx} F_{\mby}$, i.e., the joint distribution equals the product of the marginals. The statistical hypothesis is formulated as:
\begin{align*}
& H_{0}: F_{\mbx \mby}=F_{\mbx}F_{\mby},\\
& H_{A}: F_{\mbx \mby} \neq F_{\mbx}F_{\mby}.
\end{align*}
For any test statistic, the testing power at a given type $1$ error level equals the probability of correctly rejecting the null hypothesis when the random variables are dependent. A test is consistent if and only if the testing power converges to $1$ as the sample size increases to infinity, and a valid test must properly control the type $1$ error level. Modern datasets are often nonlinear, high-dimensional, and noisy, where density estimation and traditional statistical methods fail to be applicable. As multi-modal data are prevalent in much data-intensive research, a powerful, intuitive, and easy-to-use method for detecting general relationships is pivotal.

The classical Pearson's correlation \cite{Pearson1895} is still extensively employed in statistics, machine learning, and real-world applications. It is an intuitive statistic that quantifies the linear association, a special but extremely important relationship. A recent surge of interests has been placed on using distance metrics and kernel transformations to achieve consistent independence testing against all dependencies. A notable example is the distance correlation (\Dcorr) \cite{SzekelyRizzoBakirov2007, SzekelyRizzo2009, SzekelyRizzo2013a,SzekelyRizzo2014}: the population \Dcorr~is defined via the characteristic functions of the underlying random variables, while the sample \Dcorr~can be conveniently computed via the pairwise Euclidean distances of given observations. \Dcorr~enjoys universal consistency against any joint distribution of finite second moments, and is applicable to any metric space of strong negative type \cite{Lyons2013}. Notably, the idea of distance-based correlation measure can be traced back to the Mantel coefficient \cite{Mantel1967, JosseHolmes2013}: the sample version differs from sample \Dcorr~only in centering, garnered popularity in ecology and biology applications, but does not have the consistency property of \Dcorr.

Developed almost in parallel from the machine learning community, the kernel-based method (\Hsic) \cite{GrettonEtAl2005,GrettonGyorfi2010} has a striking similarity with \Dcorr: it is formulated by kernels instead of distances, can be estimated on sample data via the sample kernel matrix, and is universally consistent when using any characteristic kernel. Indeed, it is shown in \cite{SejdinovicEtAl2013} that there exists a mapping from kernel to metric (and vice versa) such that \Hsic~equals \Dcorr. Another competitive method is the Heller-Heller-Gorfine method (\Hhg) \cite{HellerGorfine2013, heller2016consistent}: it is also universally consistent by utilizing the rank information and the Pearson's chi-square test, but has better finite-sample testing powers over \Dcorr~in a collection of common nonlinear dependencies. There are other consistent methods available, such as the \Copula~method that tests independence based on the empirical copula process \cite{Genest06,Genest07,Holmes09}, entropy-based methods \cite{Mendes06}, and methods tailored for univariate data \cite{Reshef2011}.

As the number of observations in many real world problems (e.g., genetics and biology) are often limited and very costly to increase,  finite-sample testing power is crucial for certain data exploration tasks: \Dcorr~has been shown to perform well in monotone relationships, but not so well in nonlinear dependencies such as circles and parabolas; the performance of \Hsic~and \Hhg~are often the opposite of \Dcorr, which perform slightly inferior to \Dcorr~in monotone relationships but excel in various nonlinear dependencies. 

From another point of view, unraveling the nonlinear structure has been intensively studied in the manifold learning literature \cite{TenenbaumSilvaLangford2000, SaulRoweis2000, BelkinNiyogi2003}: by approximating a linear manifold locally via the k-nearest neighbors at each point, these nonlinear techniques can produce better embedding results than linear methods (like PCA) in nonlinear data. The main downside of manifold learning often lies in the parameter choice, i.e., the number of neighbor or the correct embedding dimension is often hard to estimate and requires cross-validation. Therefore, assuming a satisfactory neighborhood size can be efficiently determined in a given nonlinear relationship, the local correlation measure shall work better than the global correlation measure; and if the parameter selection is sufficiently adaptive, the optimal local correlation shall equal the global correlation in linear relationships.

In this manuscript we formalize the notion of population local distance correlations and \Mgc, explore their theoretical properties both asymptotically and in finite-sample, and propose an improved Sample \Mgc~algorithm. By combing distance correlation with the locality principle, \Mgc~inherits the universal consistency in testing, is able to efficiently search over all local scales and determine the optimal correlation, and enjoys the best testing powers throughout the simulations. A number of real data applications via \Mgc~are pursued in \cite{ShenEtAl2016}, e.g., testing brain images versus personality and disease, identify potential protein biomarkers for cancer, etc. And \Mgc~are employed for vertex dependence testing and screening in \cite{mgc3,mgc4}.

The paper is organized as follows: In Section~\ref{sec:main1}, we define the population local distance correlation and population \Mgc~via the characteristic functions of the underlying random variables and the nearest neighbor graphs, and show how the local variants are related to the distance correlation. In Section~\ref{sec:main2}, we consider the sample local correlation on finite-samples, prove its convergence to the population version, and discuss the centering and ranking scheme. In Section~\ref{sec:main3}, we present a thresholding-based algorithm for Sample \Mgc, prove its convergence property, propose a theoretically sound threshold choice, manifest that \Mgc~is valid and consistent under the permutation test, and finish the section with a number of fundamental properties for the local correlations and \Mgc. The comprehensive simulations in Section~\ref{sec:exp} exhibits the empirical advantage of \Mgc, and the paper is concluded in Section~\ref{sec:dis}.
All proofs are in Appendix~\ref{sec:proofs}, the simulation functions are presented in Appendix~\ref{appen:function}, and the code are available on Github \if1\blind
{\footnote{\url{https://github.com/neurodata/mgc-matlab}} }\fi
\if0\blind
\fi
and CRAN \if1\blind
{\footnote{\url{https://CRAN.R-project.org/package=mgc}}}\fi
\if0\blind
\fi.

\section{Multiscale Graph Correlation for Random Variables}
\label{sec:main1}

\subsection{Distance Correlation Review}
We first review the original distance correlation in \cite{SzekelyRizzoBakirov2007}. A non-negative weight function $w(t,s)$ on $(t,s) \in \mathbb{R}^{p} \times \mathbb{R}^{q}$ is defined as:
\begin{align*}
		w(t,s) &=  (c_{p}c_{q} |t|^{1+p}|s|^{1+q})^{-1},
\end{align*}
where $c_{p}=\frac{\pi^{(1+p)/2}}{\Gamma((1+p)/2)}$ is a non-negative constant tied to the dimensionality $p$, and $\Gamma(\cdot)$ is the complete Gamma function. Then the population distance covariance, variance and correlation are defined by
\begin{align*}
dCov(\mbx,\mby) &=  \int_{\mathbb{R}^{p} \times \mathbb{R}^{q}} |E(g_{\mbx \mby}(t,s))-E(g_{\mbx}(t))E(g_{\mby}(s))|^{2} w(t, s)dtds, \\
dVar(\mbx) &= dCov(\mbx,\mbx), \\
dVar(\mby) &= dCov(\mby,\mby), \\
dCorr(\mbx,\mby) &= \frac{dCov(\mbx,\mby)}{\sqrt{dVar(\mbx) \cdot dVar(\mby)}},
\end{align*}
where $|\cdot|$ is the complex modulus, $g_{\cdot}(\cdot)$ denotes the exponential transformation within the expectation of the characteristic function, i.e., $g_{\mbx \mby}(t,s) = e^{\textbf{i} \left\langle t,\mbx \right\rangle  +\textbf{i} \left\langle  s,\mby \right\rangle }$ ($\textbf{i}$ represents the imaginary unit) and $E(g_{\mbx \mby}(t,s))$ is the characteristic function. Note that distance variance equals $0$ if and only if the random variable is a constant, in which case distance correlation shall be set to $0$. The main property of population \Dcorr~is the following.
\begin{thm*}
For any two random variables $(\mbx,\mby)$ with finite first moments, $dCorr(\mbx,\mby)=0$ if and only if $\mbx$ and $\mby$ are independent. 
\end{thm*}

To estimate the population version on sample data, the sample distance covariance is computed by double centering the pairwise Euclidean distance matrix of each data, followed by summing over the entry-wise product of the two centered distance matrices. When the underlying random variables have finite second moments, the sample \Dcorr~is shown to converge to the population \Dcorr~, and is thus universally consistent for testing independence against all joint distributions of finite second moments.

\subsection{Population Local Correlations}
\label{main1a}

Next we formally define the population local distance covariance, variance, correlation by combining the k-nearest neighbor graphs with the distance covariance. For simplicity, they are named the local covariance, local variance, and local correlation from now on, and we always assume the following regularity conditions: 
\begin{align*}
& 1) \mbox{ $(\mbx,\mby)$ have finite second moments}, \\
& 2) \mbox{ Neither random variable is a constant}, \\
& 3) \mbox{ $(\mbx,\mby)$ are continuous random variables}. 
\end{align*}
The finite second moments assumption is required by \Dcorr, and also required by the local version to establish convergence and consistency. The non-constant condition is to avoid the trivial case and make sure population local correlations behave well. The continuous assumption is for ease of presentation, so the definition and related properties can be presented in a more elegant manner. Indeed, for any discrete random variable one can always apply jittering (i.e., add trivial white noise) to make it continuous without altering the independence testing. 

\begin{defi*}
Suppose $(\mbx,\mby), (\mbx',\mby'), (\mbx'',\mby''), (\mbx''',\mby''')$ are \emph{iid} as $F_{\mbx \mby}$. Let $\mb{I}(\cdot)$ be the indicator function, define two random variables
\begin{align*}
\mb{I}_{\mbx,\mbx'}^{\rho_{k}} &=\mb{I}(\int_{B(\mbx,\|\mbx'-\mbx\|)} dF_\mbx(u) \leq \rho_k)  \\
\mb{I}_{\mby',\mby}^{\rho_{l}} &=\mb{I}(\int_{B(\mby',\|\mby'-\mby\|)} dF_\mby(v) \leq \rho_l)
\end{align*}
with respect to the closed balls $B(\mbx,\|\mbx'-\mbx\|)$ and $B(\mby',\|\mby-\mby'\|)$ centered at $\mbx$ and $\mby'$ respectively. Then let $\overline{\cdot}$ denote the complex conjugate, define
\begin{align*}
h^{\rho_{k}}_{\mbx}(t) &=(g_{\mbx}(t)\overline{g_{\mbx'}(t)}-g_{\mbx}(t)\overline{g_{\mbx''}(t)}) \mb{I}_{\mbx,\mbx'}^{\rho_{k}} \\
h^{\rho_{l}}_{\mby'}(s) &=(g_{\mby'}(s)\overline{g_{\mby}(s)}-g_{\mby'}(s)\overline{g_{\mby'''}(s)}) \mb{I}_{\mby',\mby}^{\rho_{l}}
\end{align*}
as functions of $t \in \mathbb{R}^{p}$ and $s \in \mathbb{R}^{q}$ respectively,

The population local covariance, variance, correlation at any $(\rho_k,\rho_l) \in [0,1] \times [0,1]$ are defined as
\begin{align}
\label{eq:dcov1}
dCov^{\rho_k, \rho_l}(\mbx,\mby) &= \int_{\mathbb{R}^{p} \times \mathbb{R}^{q}} \{ E(h^{\rho_{k}}_{\mbx}(t) \overline{h^{\rho_{l}}_{\mby'}(s)})-E(h^{\rho_{k}}_{\mbx}(t))E(\overline{h^{\rho_{l}}_{\mby'}(s)})\} w(t, s)dtds,\\
dVar^{\rho_k}(\mbx) &= dCov^{\rho_k, \rho_k}(\mbx,\mbx), \nonumber \\
dVar^{\rho_l}(\mby) &= dCov^{\rho_l, \rho_l}(\mby,\mby), \nonumber \\
dCorr^{\rho_k,\rho_l}(\mbx,\mby) &= \frac{dCov^{\rho_k,\rho_l}(\mbx,\mby)}{\sqrt{dVar^{\rho_k}(\mbx) \cdot dVar^{\rho_l}(\mby)}},
\end{align}
where we limit the domain of population local correlation to 
\begin{align*}
\mathcal{S}_{\epsilon}=\big\{(\rho_{k},\rho_{l}) \in [0,1] \times [0,1] \mbox{ that satisfies } \min\{dVar^{\rho_k}(X),dVar^{\rho_l}(Y)\} \geq \epsilon\big\}
\end{align*}
for a small positive $\epsilon$ that is no larger than $\min\{dVar(X),dVar(Y)\}$.
\end{defi*}

The domain of local correlation needs to be limited so the population version is well-behaved. For example, when $\mbx$ is a constant or $\rho_{k}=0$, $dVar^{\rho_k}(X)$ equals $0$ and the corresponding local correlation is not well-defined. All subsequent analysis for the population local correlations is based on the domain $\mathcal{S}_{\epsilon}$, which is non-empty and compact as shown in Theorem~\ref{thmMax}. In practice, it suffices to set $\epsilon$ as any small positive number, see the sample version in Section~\ref{sec:main2}. Also note that in either indicator function, the two random variables and the distribution $dF$ are independent, e.g., at any realization $(x, x')$ of $(\mbx,\mbx')$, the first indicator equals $\mb{I}(\int_{B(x,\|x'-x\|)} dF_\mbx(u) \leq \rho_k)$, and its expectation is taken with respect to $(\mbx,\mbx')$. 

The above definition makes use of the characteristic functions, which is akin to the original definition of \Dcorr~and easier to show consistency. Alternatively, the local covariance can be equivalently defined via the pairwise Euclidean distances. The alternative definition better motivates the sample version in Section~\ref{sec:main2}, is often handy for understanding and proving theoretical properties, and suggests that local covariance is always a real number, which is not directly obvious from Equation~\ref{eq:dcov1}.
\begin{thm}
\label{thm1}
Suppose $(\mbx,\mby),(\mbx',\mby'),(\mbx'',\mby''),(\mbx''',\mby''')$ are \emph{iid} as $F_{\mbx \mby}$, and define
\begin{align*}
d^{\rho_{k}}_{\mbx} &=(\| \mbx-\mbx' \| - \|\mbx-\mbx''\|) \mb{I}_{\mbx,\mbx'}^{\rho_{k}} \\
d^{\rho_{l}}_{\mby'} &=(\| \mby'-\mby \| - \|\mby'-\mby'''\|) \mb{I}_{\mby',\mby}^{\rho_{l}}
\end{align*}

The local covariance in Equation~\ref{eq:dcov1} can be equally defined as
\begin{align}
\label{eq:dcov2}
dCov^{\rho_k, \rho_l}(\mbx,\mby) = E(d^{\rho_{k}}_{\mbx} d^{\rho_{l}}_{\mby'}) - E(d^{\rho_{k}}_{\mbx}) E(d^{\rho_{l}}_{\mby'}),
\end{align}
which shows that local covariance, variance, correlation are always real numbers.
\end{thm} 

Each local covariance is essentially a local version of distance covariance that truncates large distances at each point in the support, where the neighborhood size is determined by $(\rho_{k},\rho_{l})$. In particular, distance correlation equals the local correlation at the maximal scale, which will ensure the consistency of \Mgc.

\begin{thm}
\label{thm2}
At any $(\rho_k,\rho_l)\in \mathcal{S}_{\epsilon}$, $dCov^{\rho_{k},\rho_{l}}(\mbx,\mby) = 0$ when $\mbx$ and $\mby$ are independent. Moreover, at $(\rho_k,\rho_l)=(1,1)$, $dCov^{\rho_{k},\rho_{l}}(\mbx,\mby) = dCov(\mbx,\mby)$. They also hold for the correlations by replacing all the $dCov$ by $dCorr$.
\end{thm}

\subsection{Population \Mgc~and Optimal Scale}
The population \Mgc~can be naturally defined as the maximum local correlation within the domain, i.e., 
\begin{align}
\label{eq:pmgc}
\GG^{*}(\mbx,\mby)=\max_{ (\rho_k,\rho_l) \in \mathcal{S}_{\epsilon}} \{dCorr^{\rho_{k},\rho_{l}}(\mbx,\mby)\},
\end{align}
and the scale that attains the maximum is named the optimal scale
\begin{align}
\label{eq:pmgc1}
(\rho_k,\rho_l)^{*}=\arg\max_{ (\rho_k,\rho_l) \in \mathcal{S}_{\epsilon}} \{dCorr^{\rho_{k},\rho_{l}}(\mbx,\mby)\}.
\end{align}
The next theorem states the continuity of the local covariance, variance, correlation, and thus the existence of population \Mgc.
\begin{thm}
\label{thmMax}
Given two continuous random variables $(\mbx,\mby)$, 
\begin{description}
\item [(a)] The local covariance is a continuous function with respect to $(\rho_{k},\rho_{l}) \in [0,1]^2$, so is local variance in $[0,1]$ and local correlation in $\mathcal{S}_{\epsilon}$. 
\item [(b)] The set $\mathcal{S}_{\epsilon}$ is always non-empty unless either random variable is a constant.
\item [(c)] Excluding the trivial case in (b), the set $\{dCorr^{\rho_{k},\rho_{l}}(\mbx,\mby), (\rho_k,\rho_l) \in \mathcal{S}_{\epsilon}\}$ is always non-empty and compact, so an optimal scale $(\rho_k,\rho_l)^{*}$ and $\GG^{*}(\mbx,\mby)$ exist.
\end{description}
\end{thm} 

Therefore, population \Mgc~and the optimal scale exist, are distribution dependent, and may not be unique. Without loss of generality, the optimal scale is assumed unique for presentation purpose. The population \Mgc~is always no smaller than \Dcorr~in magnitude, and equals $0$ if and only if independence, a property inherited from \Dcorr.
\begin{thm}
\label{thm3}
When $\mbx$ and $\mby$ are independent, $\GG^{*}(\mbx,\mby)=dCorr(\mbx,\mby)=0$; when $\mbx$ and $\mby$ are not independent, $\GG^{*}(\mbx,\mby) \geq dCorr(\mbx,\mby)>0$.
\end{thm} 

\section{Sample Local Correlations}
\label{sec:main2}

Sample \Dcorr~can be easily calculated via properly centering the Euclidean distance matrices, and is shown to converge to the population \Dcorr~\cite{SzekelyRizzoBakirov2007, SzekelyRizzo2013a, SzekelyRizzo2014}. Similarly, we show that the sample local correlation can be calculated via the Euclidean distance matrices upon truncating large distances for each sample observation, and the sample version converges to the respective population local correlation.

\subsection{Definition}
\label{sec:defi}
Given pairs of observations $(x_{i},y_{i}) \in \Real^{p} \times \Real^{q}$ for $i=1,\ldots,n$, denote $\mathcal{X}_{n}=[x_{1},\ldots,x_{n}]$ as the data matrix with each column representing one sample observation, and similarly $\mathcal{Y}_{n}$. Let $\tilde{A}$ and $\tilde{B}$ be the $n \times n$ Euclidean distance matrices of $\mathcal{X}_{n}=\{x_{i}\}$ and $\mathcal{Y}_{n}=\{y_{i}\}$ respectively, i.e., $\tilde{A}_{ij}=\|x_{i}-x_{j}\|$. Then we compute two column-centered matrices $A$ and $B$ with the diagonals excluded, i.e., $\tilde{A}$ and $\tilde{B}$ are centered within each column such that
\begin{equation}
\label{localCoef2}
    A_{ij}=
    \begin{cases}
      \tilde{A}_{ij}-\frac{1}{n-1}\sum_{s=1}^{n} \tilde{A}_{sj}, & \text{if $i \neq j$}, \\    
      0, & \text{if $i=j$};
    \end{cases} \qquad \qquad
    B_{ij}=
    \begin{cases}
      \tilde{B}_{ij}-\frac{1}{n-1}\sum_{s=1}^{n} \tilde{B}_{sj}, & \text{if $i \neq j$}, \\    
      0, & \text{if $i=j$};
    \end{cases}
\end{equation}

Next we define $\{R^{A}_{ij}\}$ as the ``rank'' of $x_i$ relative to $x_j$, that is, $R^{A}_{ij}=k$ if $x_i$ is the $k^{th}$ closest point (or ``neighbor'') to $x_j$, as determined by ranking the set $\{\tilde{A}_{1j},\tilde{A}_{2j},\ldots,\tilde{A}_{nj}\}$ by ascending order. Similarly define $R^{B}_{ij}$ for the $y$'s. As we assumed $(\mbx,\mby)$ are continuous, with probability $1$ there is no repeating observation and the ranks always take value in $\{1,\ldots,n\}$. In practice ties may occur, and we recommend either using minimal rank to keep the ties or jittering to break the ties, which is discussed at the end of this section.

For any $(k,l) \in [n]^2=\{1,\ldots,n\} \times \{1,\ldots,n\}$, we define the rank truncated matrices $A^{k}$ and $B^{l}$ as
\begin{align*}
A_{ij}^{k} &=A_{ij} \mb{I}(R^{A}_{ij} \leq k), \\
B_{ij}^{l} &=B_{ij} \mb{I}(R^{B}_{ij} \leq l).
\end{align*}
Let $\circ$ denote the entry-wise product, $\E(\cdot)=\frac{1}{n(n-1)}\sum_{i \neq j}^{n} (\cdot)$ denote the diagonal-excluded sample mean of a square matrix, then the sample local covariance, variance, and correlation are defined as:
\begin{align*}
dCov^{k,l}(\mathcal{X}_{n},\mathcal{Y}_{n}) &= \E(A^{k} \circ B^{l'})- \E(A^{k})\E(B^{l}),\\
dVar^{k}(\mathcal{X}_{n}) &=\E(A^{k} \circ A^{k'})- \E^2(A^{k}), \\
dVar^{l}(\mathcal{Y}_{n}) &=\E(B^{l} \circ B^{l'})- \E^2(B^{l}), \\
dCorr^{k,l}(\mathcal{X}_{n},\mathcal{Y}_{n}) &=dCov^{k,l}(\mathcal{X}_{n},\mathcal{Y}_{n}) / \sqrt{dVar^{k}(\mathcal{X}_{n}) \cdot dVar^{l}(\mathcal{Y}_{n})}.
\end{align*}
If either local variance is smaller than a preset $\epsilon > 0$ (e.g., the smallest positive local variance among all), then we set the corresponding $dCorr^{k,l}(\mathcal{X}_{n},\mathcal{Y}_{n})=0$ instead. Note that once the rank is known, sample local correlations can be iteratively computed in $\mathcal{O}(n^2)$ rather than a naive implementation of $\mathcal{O}(n^3)$. A detailed running time comparison is presented in Section~\ref{sec:exp}.

In case of ties, minimal rank offers a consecutive indexing of sample local correlations, e.g., if $\mby$ only takes two values, $R^{B}_{ij}$ takes value in $\{1,2\}$ under minimal rank, but maximal rank yields $\{\frac{n}{2},n\}$. The sample local correlations are not affected by the tie scheme, but minimal rank is more convenient to work with for implementation purposes. Alternatively, one can break ties deterministically or randomly, e.g., apply jittering to break all ties. For example, in the Bernoulli relationship of Figure~\ref{f:dependencies}, there are only three points for computing sample local correlations and the Sample \Mgc~equals $0.9$. If white noise of variance $0.01$ were added to the data, we break all ties and obtain a much larger number of sample local correlations. The resulting Sample \Mgc~is $0.8$, which is slightly smaller but still much larger than $0$ and implies a strong dependency.

Whether the random variable is continuous or discrete, and whether the ties in sample data are broken or not, does not affect the theoretical results except in certain theorem statements. For example, in Theorem~\ref{thm4}, the convergence still holds for discrete random variables, but the index pair $(k,l)$ does not necessarily correspond to the population version at $(\rho_{k},\rho_{l})=(\frac{k-1}{n-1}, \frac{l-1}{n-1})$, e.g., when $\mbx$ is Bernoulli with probability $0.8$ and minimal rank is used, $k=1$ corresponds to $\rho_{k}=0.8$ instead of $\rho_{k}=\frac{k-1}{n-1}$. Nevertheless, Theorem~\ref{thm4} and all results in the paper hold regardless of continuous or discrete random variables, but the presentation is more elegant for the continuous case.

\subsection{Convergence Property}
The sample local covariance, variance, correlation are designed to converge to the respective population versions. Moreover, the expectation of sample local covariance equals the population counterpart up to a difference of $\mathcal{O}(\frac{1}{n})$, and the variance diminishes at the rate of $\mathcal{O}(\frac{1}{n})$.
\begin{thm}
\label{thm4}
Suppose each column of $\mathcal{X}_{n}$ and $\mathcal{Y}_{n}$ are generated \emph{iid} from $(\mbx,\mby) \sim F_{\mbx \mby}$. The sample local covariance satisfies
\begin{align*}
E(dCov^{k,l}(\mathcal{X}_{n},\mathcal{Y}_{n})) &= dCov^{\rho_{k},\rho_{l}}(\mbx,\mby) +\mathcal{O}(1/n) \\
Var(dCov^{k,l}(\mathcal{X}_{n},\mathcal{Y}_{n})) &= \mathcal{O}(1/n)\\
dCov^{k,l}(\mathcal{X}_{n},\mathcal{Y}_{n}) &\stackrel{n \rightarrow \infty}{\rightarrow} dCov^{\rho_{k},\rho_{l}}(\mbx,\mby),
\end{align*}
where $\rho_{k}=\frac{k-1}{n-1}$ and $\rho_{l}=\frac{l-1}{n-1}$. In particular, the convergence is uniform and also holds for the local correlation, i.e., for any $\epsilon$ there exists $n_{\epsilon}$ such that for all $n > n_{\epsilon}$, 
\begin{align*}
|dCorr^{k,l}(\mathcal{X}_{n},\mathcal{Y}_{n}) -dCorr^{\rho_{k},\rho_{l}}(\mbx,\mby)|< \epsilon
\end{align*}
for any pair of $(\rho_{k},\rho_{l}) \in \mathcal{S}_{\epsilon}$.
\end{thm}

The convergence property ensures that Theorem~\ref{thm2} holds asymptotically for the sample version.
\begin{cor}
\label{thm5}
For any $(k,l)$, $dCorr^{k,l}(\mathcal{X}_{n},\mathcal{Y}_{n}) \rightarrow 0$ when $\mbx$ and $\mby$ are independent. In particular, $dCorr^{n,n}(\mathcal{X}_{n},\mathcal{Y}_{n}) \rightarrow dCorr(\mbx,\mby)$.

\end{cor}

Moreover, one can show that $dCorr^{n,n}(\mathcal{X}_{n},\mathcal{Y}_{n}) \approx dCorr(\mathcal{X}_{n},\mathcal{Y}_{n})$ for the unbiased sample distance correlation in \cite{SzekelyRizzo2014} up-to a small difference of $\mathcal{O}(\frac{1}{n})$, which can be verified by comparing Equation~\ref{localCoef2} to Equation 3.1 in \cite{SzekelyRizzo2014}. 

\subsection{Centering and Ranking}
To combine distance testing with the locality principle, other than the procedure proposed in Equation~\ref{eq:dcov2}, there are a number of alternative options to center and rank the distance matrices. For example, letting
\begin{align*}
d^{\rho_{k}}_{\mbx} &=(\| \mbx-\mbx' \| - \|\mbx-\mbx''\|-\|\mbx'-\mbx''\|+\|\mbx''-\mbx'''\|) \mb{I}_{\mbx,\mbx'}^{\rho_{k}}, \\
d^{\rho_{l}}_{\mby'} &=(\| \mby'-\mby \| - \|\mby'-\mby''\|-\|\mby-\mby''\|+\|\mby''-\mby'''\|) \mb{I}_{\mby',\mby}^{\rho_{l}}
\end{align*}
still guarantees the resulting local correlation at maximal scale equals the distance correlation; and letting 
\begin{align*}
d^{\rho_{k}}_{\mbx} &=\| \mbx-\mbx' \|  \mb{I}_{\mbx,\mbx'}^{\rho_{k}}, \\
d^{\rho_{l}}_{\mby'} &=\| \mby'-\mby \|  \mb{I}_{\mby',\mby}^{\rho_{l}},
\end{align*}
makes the resulting local correlation at maximal scale equal the \Mantel~coefficient, the earliest distance-based correlation coefficient. 

Nevertheless, the centering and ranking strategy proposed in Equation~\ref{eq:dcov2} is more faithful to k-nearest neighbor graph: the indicator $\mb{I}_{\mbx,\mbx'}^{\rho_{k}}$ equals $1$ if and only if $\int_{B(\mbx,\|\mbx'-\mbx\|)} dF_\mbx(u) \leq \rho_k$, which happens with probability $\rho_{k}$. Viewed another way, when conditioned on $(\mbx,\mbx')=(x,x')$, the indicator equals $1$ if and only if $Prob(\|x'-x\|<\|\mbx''-x\|) \leq \rho_{k}$, thus matching the column ranking scheme in Equation~\ref{localCoef2}. Indeed, the locality principle used in \cite{TenenbaumSilvaLangford2000, SaulRoweis2000, BelkinNiyogi2003} considers the k-nearest neighbors of each sample point in local computation, an essential step to yield better nonlinear embeddings. 

On the centering side, the \Mantel~test appears to be an attractive option due to its simplicity in centering. All the \Dcorr, \Hhg, \Hsic~have their theoretical consistency, while the \Mantel~coefficient does not, despite it being merely a different centering of \Dcorr. An investigation of the population form of \Mantel~yields some additional insights:
\begin{defi*}
Given $\mathcal{X}_{n}$ and $\mathcal{Y}_{n}$, the \Mantel~coefficient on sample data is computed as 
\begin{align*}
M(\mathcal{X}_{n},\mathcal{Y}_{n}) &= \E(\tilde{A} \circ \tilde{B})-\E(\tilde{A})\E(\tilde{B}) \\
Mantel(\mathcal{X}_{n},\mathcal{Y}_{n}) &= \frac{M(\mathcal{X}_{n},\mathcal{Y}_{n})}{\sqrt{M(\mathcal{X}_{n},\mathcal{X}_{n}) M(\mathcal{Y}_{n},\mathcal{Y}_{n})}},
\end{align*}
where $\tilde{A}_{ij}$ and $\tilde{B}_{ij}$ are the pairwise Euclidean distance, and $\E(\cdot)=\frac{1}{n(n-1)}\sum_{i \neq j}^{n} (\cdot)$ is the diagonal-excluded sample mean of a square matrix.
\end{defi*}

\begin{cor}
\label{cor1}
Suppose each column of $\mathcal{X}_{n}$ and $\mathcal{Y}_{n}$ are \emph{iid} as $F_{\mbx \mby}$, and $(\mbx,\mby), (\mbx',\mby')$ are also \emph{iid} as $F_{\mbx \mby}$. Then 
\begin{align*}
Mantel(\mathcal{X}_{n},\mathcal{Y}_{n}) &\rightarrow Mantel(\mbx,\mby) = \frac{M(\mbx,\mby)}{\sqrt{M(\mbx,\mbx) M(\mby,\mby)}},
\end{align*}
where
\begin{align*}
M(\mbx,\mby) &= \int_{\mathbb{R}^{p} \times \mathbb{R}^{q}} \{|E(g_{\mbx \mby}(t,s))|^2-|E(g_{\mbx}(t))E(g_{\mby}(s))|^{2}\} w(t, s)dtds \\
&= E(\| \mbx-\mbx' \| \| \mby-\mby' \|) - E(\|\mbx-\mbx'\|)E(\|\mby-\mby'\|)) \\
&= Cov(\| \mbx-\mbx' \|, \| \mby-\mby' \|).
\end{align*}
\end{cor}
Corollary~\ref{cor1} suggests that \Mantel~is actually a two-sided test based on the absolute difference of characteristic functions: under certain dependency structure, the \Mantel~coefficient can be negative and still imply dependency (i.e., $|E(g_{\mbx \mby}(t,s))| < |E(g_{\mbx}(t))E(g_{\mby}(s))|$); whereas population \Dcorr~and \Mgc~are always no smaller than $0$, and any negativity of the sample version does not imply dependency. Therefore, \Mantel~is only appropriate as a two-sided test, which is evaluated in Section~\ref{sec:exp}. 

Another insight is that \Mantel, unlike \Dcorr, is not universally consistent: due to the integral $w$, one can construct a joint distribution such that the population \Mantel~equals $0$ under dependence (see Remark 3.13 in \cite{Lyons2013} for an example of dependent random variables with uncorrelated distances). However, empirically, simple centering is still effective in a number of common dependencies (like two parabolas and diamond in Figure~\ref{f:1DAll}). 

\section{Sample \Mgc~and Estimated Optimal Scale}
\label{sec:main3}

A naive sample version of \Mgc~can be defined as the maximum of all sample local correlations 
\begin{align*}
\max_{(k,l) \in [n]^2}\{dCorr^{k,l}(\mathcal{X}_{n},\mathcal{Y}_{n})\}.
\end{align*}
Although the convergence to population \Mgc~can be guaranteed, the sample maximum is a biased estimator of the population \Mgc~in Equation~\ref{eq:pmgc}. For example, under independence, population \Mgc~equals $0$, while the maximum sample local correlation has expectation larger than $0$, which may negate the advantage of searching locally and hurt the testing power. 

This motivates us to compute Sample \Mgc~as a smoothed maximum within the largest connected region of thresholded local correlations. The purpose is to mitigate the bias of a direct maximum, while maintaining its advantage over \Dcorr~in the test statistic. The idea is that in case of dependence, local correlations on the grid near the optimal scale shall all have large correlations; while in case of independence, a few local correlations may happen to be large, but most nearby local correlations shall still be small. The idea can be similarly adapted whenever there are multiple correlated test statistics or multiple models available, for which taking a direct maximum may yield too much bias \cite{mgc3}. From another perspective, Sample \Mgc~is like taking a regularized maximum.

\subsection{Sample \Mgc}
The procedure is as follows:
\begin{description}[align=left]
\item [Input: ] A pair of datasets $(\mathcal{X}_{n}, \mathcal{Y}_{n})$.
\item [Compute the Local Correlation Map:] Compute all local correlations:\\
$\{dCorr^{k,l}(\mathcal{X}_{n},\mathcal{Y}_{n}), (k,l) \in [n]^2\}$.
\item [Thresholding: ] Pick a threshold $\tau_n \geq 0$, denote $LC(\cdot)$ as the operation of taking the largest connected component, and compute the largest region $R$ of thresholded local correlations:
\begin{align}
\label{eq:region}
&R=LC(\{(k,l) \mbox{ such that } dCorr^{k,l}(\mathcal{X}_{n},\mathcal{Y}_{n})>\max\{\tau_n, dCorr^{n,n}(\mathcal{X}_{n},\mathcal{Y}_{n})\} \}). 
\end{align}
Within the region $R$, set
\begin{align}
\label{eq:smgc1}
\GG^{*}(\mathcal{X}_{n},\mathcal{Y}_{n})&=\max_{ (k,l) \in R} \{dCorr^{k,l}(\mathcal{X}_{n},\mathcal{Y}_{n})\}\\
(k_{n},l_{n})^{*}&=\arg\max_{ (k,l) \in R} \{dCorr^{k,l}(\mathcal{X}_{n},\mathcal{Y}_{n})\}
\end{align}
as the Sample \Mgc~and the estimated optimal scale. If the number of elements in $R$ is less than $2n$, or the above thresholded maximum is no more than $dCorr^{n,n}(\mathcal{X}_{n},\mathcal{Y}_{n})$, we instead set $\GG^{*}(\mathcal{X}_{n},\mathcal{Y}_{n})=dCorr^{n,n}(\mathcal{X}_{n},\mathcal{Y}_{n})$ and $(k_{n},l_{n})^{*}=(n,n)$.

\item [Output: ] Sample \Mgc~$\GG^{*}(\mathcal{X}_{n}, \mathcal{Y}_{n})$ and the estimated optimal scale $(k_{n},l_{n})^{*}$.
\end{description}

If there are multiple largest regions, e.g., $R_{1}$ and $R_{2}$ where their number of elements are more than $2n$ and coincide with each other, then it suffices to let $R = R_{1} \displaystyle \cup R_{2}$ and locate the \Mgc~statistic within the union. The selection of at least $2n$ elements for $R$ is an empirical choice, which balances the bias-variance trade-off well in practice. The parameter can be any positive integer without affecting the validity and consistency of the test. But if the parameter is too large, \Mgc~tends to be more conservative and is unable to detect signals in strongly nonlinear relationships (e.g., trigonometric functions), and performs closer and closer to \Dcorr; if the parameter is set to a very small fixed number, the bias is inflated so \Mgc~tends to perform similarly as directly maximizing all local correlations.

\subsection{Convergence and Consistency}
The proposed Sample \Mgc~is algorithmically enforced to be no less than the local correlation at the maximal scale, and also no more than the maximum local correlation. It also ensures in Theorem~\ref{thm3} to hold for the sample version.
\begin{thm}
\label{thm6}
Regardless of the threshold $\tau_n$, the Sample \Mgc~statistic $\GG^{*}(\mathcal{X}_{n},\mathcal{Y}_{n})$ satisfies
\begin{description}
\item [(a)] It always holds that
\begin{align*}
\max_{(k,l) \in [n]^2}\{dCorr^{k,l}(\mathcal{X}_{n},\mathcal{Y}_{n})\} \geq \GG^{*}(\mathcal{X}_{n},\mathcal{Y}_{n}) \geq dCorr^{n,n}(\mathcal{X}_{n},\mathcal{Y}_{n}).
\end{align*}
\item [(b)] When $\mbx$ and $\mby$ are independent, $\GG^{*}(\mathcal{X}_{n},\mathcal{Y}_{n}) \rightarrow 0$; when $\mbx$ and $\mby$ are not independent, $\GG^{*}(\mathcal{X}_{n},\mathcal{Y}_{n}) \rightarrow$ a positive constant.
\end{description}
\end{thm}

The next theorem states that if the threshold $\tau_n$ converges to $0$, then whenever population \Mgc~is larger than population \Dcorr, Sample \Mgc~is also larger than sample \Dcorr~asymptotically; otherwise if the threshold does not converge to $0$, Sample \Mgc~may equal sample \Dcorr~despite of the first moment advantage in population. Moreover, Sample \Mgc~indeed converges to population \Mgc~when the optimal scale is in the largest thresholded region $R$. The empirical advantage of Sample \Mgc~is illustrated in Figure~\ref{f:dependencies}. 

\begin{thm}
\label{thm7}
Suppose each column of $\mathcal{X}_{n}$ and $\mathcal{Y}_{n}$ are \emph{iid} as continuous $(\mbx, \mby) \sim F_{\mbx \mby}$, and the threshold choice $\tau_n \rightarrow 0$ as $n \rightarrow \infty$.
\begin{description}
\item [(a)] Assume that $\GG^{*}(\mbx, \mby) > Dcorr(\mbx, \mby)$ under the joint distribution. Then $\GG^{*}(\mathcal{X}_{n}, \mathcal{Y}_{n}) > Dcorr(\mathcal{X}_{n}, \mathcal{Y}_{n})$ for $n$ sufficiently large. 
\item [(b)] Assume there exists an element within the the largest connected area of $\{(\rho_k,\rho_l) \in \mathcal{S}_{\epsilon}$ with $dCorr^{\rho_k,\rho_l}(\mbx,\mby)> dCorr(\mbx,\mby) \}$, such that the the local correlation of that element equals $\GG^{*}(\mbx,\mby)$. Then $\GG^{*}(\mathcal{X}_{n}, \mathcal{Y}_{n}) \rightarrow \GG^{*}(\mbx,\mby)$.
\end{description}
\end{thm}
Alternatively, Theorem~\ref{thm7}(b) can be stated that the Sample \Mgc~always converges to the maximal population local correlation within the largest connected area of thresholded local correlations. Therefore, Sample \Mgc~converges either to \Dcorr~(when the area is empty) or something larger, thus improving over \Dcorr~statistic in first moment.

\subsection{Choice of Threshold}

The choice of threshold $\tau_n$ is imperative for Sample \Mgc~to enjoy a good finite-sample performance, especially at small sample size. According to Theorem~\ref{thm7}, the threshold shall converge to $0$ for Sample \Mgc~to prevail sample \Dcorr.

A model-free threshold $\tau_n$ was previously used in \cite{ShenEtAl2016}: for the following set 
\begin{align*}
\{dCorr^{k,l}(\mathcal{X}_{n},\mathcal{Y}_{n}) \mbox{ s.t. }  dCorr^{k,l}(\mathcal{X}_{n},\mathcal{Y}_{n})<0\},
\end{align*}
let $\sigma^2$ be the sum of all its elements squared, and set $\tau_n=5\sigma$ as the threshold; if there is no negative local correlation and the set is empty, use $\tau_n=0.05$. 

Although the previous threshold is a data-adaptive choice that works pretty well empirically and does not affect the consistency of Sample \Mgc~in Theorem~\ref{thm8}, it does not converge to $0$. The following finite-sample theorem from \cite{SzekelyRizzo2013a} motivates an improved threshold choice here:
\begin{thm*}
Under independence of $(\mbx,\mby)$, assume the dimensions of $\mbx$ are exchangeable with finite variance, and so are the dimensions of $\mby$. Then for any $n \geq 4$ and $v=\frac{n(n-3)}{2}$, as $p,q$ increase the limiting distribution of $(dCorr^{n,n}(\mathcal{X}_{n}, \mathcal{Y}_{n})+1)/2$ equals the symmetric Beta distribution with shape parameter $\frac{v-1}{2}$.
\end{thm*}

The above theorem leads to the new threshold choice:
\begin{cor}
\label{cor2}
Denote $v=\frac{n(n-3)}{2}$, $z \sim Beta(\frac{v-1}{2})$, $F^{-1}_{z}(\cdot)$ as the inverse cumulative distribution function. The threshold choice
\begin{align*}
\tau_n= 2F^{-1}_{z} \Big(1-\frac{0.02}{n}\Big)-1 
\end{align*} 
converges to $0$ as $n \rightarrow \infty$.
\end{cor}
The limiting null distribution of \Dcorr~is still a good approximation even when $p,q$ are not large, thus provides a reliable bound for eliminating local correlations that are larger than \Dcorr~by chance or by noise. The intuition is that Sample \Mgc~is mostly useful when it is much larger than \Dcorr~in magnitude, which is often the case in non-monotone relationships as shown in Section~\ref{sec:exp} Figure~\ref{f:dependencies}. Alternatively, directly setting $\tau_{n}=0$ also guarantees the theoretical properties and works equally well when the sample size $n$ is moderately large.

\subsection{Permutation Test}
\label{sec:permutation}

To test independence on a pair of sample data $(\mathcal{X}_{n},\mathcal{Y}_{n})$, the random permutation test has been the popular choice \cite{GoodPermutationBook} for almost all methods introduced, as the null distribution of the test statistic can be easily approximated by randomly permuting one data set. We discuss the computation procedure, prove the testing consistency of \Mgc, and analyze the running time.

To compute the p-value of \Mgc~from the permutation test, first compute the Sample \Mgc~statistic $\GG^{*}(\mathcal{X}_{n},\mathcal{Y}_{n})$ on the observed data pair. Then the \Mgc~statistic is repeatedly computed on the permuted data pair, e.g. $\mathcal{Y}_{n}=[y_{1},\ldots,y_{n}]$ is permuted into $\mathcal{Y}_{n}^{\pi}=[y_{\pi(1)},\ldots,y_{\pi(n)}]$ for a random permutation $\pi$ of size $n$, and compute $\GG^{*}(\mathcal{X}_{n},\mathcal{Y}_{n}^{\pi})$. The permutation procedure is repeated for $r$ times to estimate the probability $Prob(\GG^{*}(\mathcal{X}_{n},\mathcal{Y}_{n}^{\pi}) > \GG^{*}(\mathcal{X}_{n},\mathcal{Y}_{n}))$, and the estimated probability is taken as the p-value of \Mgc. The independence hypothesis is rejected if the p-value is smaller than a pre-set critical level, say $0.05$ or $0.01$. The following theorem states that \Mgc~via the permutation test is consistent and valid.

\begin{thm}
\label{thm8}
Suppose each column of $\mathcal{X}_{n}$ and $\mathcal{Y}_{n}$ are generated \emph{iid} from $F_{\mbx \mby}$. At any type $1$ error level $\alpha>0$, Sample \Mgc~is a valid test statistic that is consistent against all possible alternatives under the permutation test.
\end{thm}

\subsection{Miscellaneous Properties}
\label{sec:misc}
  
In this subsection, we first show a useful lemma expressing sample local covariance in Section~\ref{sec:defi} by matrix trace and eigenvalues, then list a number of fundamental and desirable properties for the local variance, local correlation, and \Mgc, akin to these of Pearson's correlation and distance correlation as shown in \cite{SzekelyRizzoBakirov2007, SzekelyRizzo2009}.

\begin{lem}
\label{lem1}
Denote $tr(\cdot)$ as the matrix trace, $\lambda_{i} [\cdot]$ as the $i$th eigenvalue of a matrix, and $J$ as the matrix of ones of size $n$. Then the sample covariance equals
\begin{align*}
dCov^{k,l}(\mathcal{X}_{n},\mathcal{Y}_{n}) &= tr (A^{k}B^{l})- tr (A^{k}J)tr(B^{l}J)\\
&=tr [ (A^{k}- tr (A^{k}J)J) (B^{l}-tr(B^{l}J)J)] \\
& = \sum_{i=1}^{n} \lambda_{i} [ (A^{k}- tr (A^{k}J)J) (B^{l}-tr(B^{l}J)J)].
\end{align*}
\end{lem}

\begin{thm}[Local Variances]
\label{thm:dvar}
For any random variable $\mbx \sim F_{\mbx} \in \mathbb{R}^{p}$, and any $\mathcal{X}_{n} \in \mathbb{R}^{p \times n}$ with each column \emph{iid} as $F_{\mbx}$,
\begin{description}
\item[(a)] Population and sample local variances are always non-negative, i.e., 
\begin{align*}
dVar^{\rho_{k}}(\mbx) \geq 0\\
dVar^{k}(\mathcal{X}_{n}) \geq 0
\end{align*}
at any $\rho_{k} \in [0,1]$ and  any $k \in [n]$.
\item[(b)] $dVar^{\rho_{k}}(\mbx) =0$ if and only if either $\rho_k =0$ or $F_{\mbx}$ is a degenerate distribution; 

$dVar^{k}(\mathcal{X}_{n}) =0$ if and only if either $k=1$ or $F_{\mbx}$ is a degenerate distribution.

\item[(c)] For two constants $v \in \mathbb{R}^{p},u \in \mathbb{R}$, and an orthonormal matrix $Q \in \mathbb{R}^{p \times p}$, 
\begin{align*}
dVar^{\rho_{k}}(v+uQ\mbx) &=u^2 \cdot dVar^{\rho_{k}}(\mbx)\\
dVar^{k}(v^{T} J +u\mathcal{X}_{n} Q) &=u^2 \cdot dVar^{k}(\mathcal{X}_{n} ).
\end{align*}
\end{description}
\end{thm}

Therefore, the local variances end up having properties similar to the distance variance in \cite{SzekelyRizzoBakirov2007}, except the distance variance definition there takes a square root.

\begin{thm}[Local Correlations and \Mgc]
\label{thm:dcor}
For any pair of random variable $(\mbx,\mby) \sim F_{\mbx\mby} \in \mathbb{R}^{p} \times \mathbb{R}^{q}$, and any $(\mathcal{X}_{n},\mathcal{Y}_{n}) \in \mathbb{R}^{p \times n} \times \mathbb{R}^{q \times n}$ with each column \emph{iid} as $F_{\mbx\mby}$,
\begin{description}
\item [(a)] Symmetric and Boundedness:
\begin{align*}
dCorr^{\rho_{k},\rho_{l}}(\mbx,\mby)=dCorr^{\rho_{l},\rho_{k}}(\mby,\mbx) \in [-1,1]\\
dCorr^{k,l}(\mathcal{X}_{n},\mathcal{Y}_{n})=dCorr^{l,k}(\mathcal{Y}_{n},\mathcal{X}_{n}) \in [-1,1]
\end{align*}
at any $(\rho_k,\rho_l) \in (0,1]^2$ and any $(k,l) \in [2,\ldots,n]^2$.
\item [(b)] Assume $F_\mbx$ is non-degenerate. Then at any $\rho_{k} > 0$, $dCorr^{\rho_{k},\rho_k}(\mbx,\mby)=1$ if and only if $(\mbx, u \mby)$ are dependent via an isometry for some non-zero constant $u \in \mathbb{R}$. 

Assume $F_\mbx$ is non-degenerate. Then at any $k > 1$, $dCorr^{k,k}(\mathcal{X}_{n},\mathcal{Y}_{n})=1$ if and only if $(\mbx, u \mby)$ are dependent via an isometry for some non-zero constant $u \in \mathbb{R}$.

\item [(c)] Both population and Sample \Mgc~are symmetric and bounded:
\begin{align*}
\GG^{*}(\mbx,\mby)=\GG^{*}(\mby,\mbx) \in [-1,1] \\
\GG^{*}(\mathcal{X}_{n},\mathcal{Y}_{n})=\GG^{*}(\mathcal{Y}_{n},\mathcal{X}_{n}) \in [-1,1].
\end{align*}
\item [(d)] Assume $F_\mbx$ is non-degenerate. Then $\GG^{*}(\mbx,\mby)=1$ if and only if $(\mbx, u \mby)$ are dependent via an isometry for some non-zero constant $u \in \mathbb{R}$. 

Assume $F_\mbx$ is non-degenerate. Then $\GG^{*}(\mathcal{X}_{n},\mathcal{Y}_{n})=1$ if and only if $(\mbx, u \mby)$ are dependent via an isometry for some non-zero constant $u \in \mathbb{R}$.
\end{description}
\end{thm}

The proof of Theorem~\ref{thm:dcor}(b)(d) also shows that the local correlations and \Mgc~cannot be $-1$.

\section{Experiments}
\label{sec:exp}
In the experiments, we compare Sample \Mgc~with \Dcorr, \Pearson, \Mantel, \Hsic, \Hhg, and \Copula~test on $20$ different simulation settings based on a combination of simulations used in previous works \cite{SzekelyRizzoBakirov2007, SimonTibshirani2012, GorfineHellerHeller2012}. Among the $20$ settings, the first $5$ are monotonic relationships (and several of them are linear or nearly so), the last simulation is an independent relationship, and the remaining settings consist of common non-monotonic and strongly nonlinear relationships. The exact distributions are shown in Appendix. 

\subsection*{The Sample Statistics}

Figure~\ref{f:dependencies} shows the sample statistics of \Mgc, \Dcorr, and \Pearson~for each of the $20$ simulations in a univariate setting. For each simulation, we generate sample data $(\mathcal{X}_{n},\mathcal{Y}_{n})$ at $p=q=1$ and $n=100$ without any noise, then compute the sample statistics. From type $1-5$, the test statistics for both \Mgc~and \Dcorr~are remarkably greater than $0$ and almost identical to each other. For the nonlinear relationships  (type $6-19$), \Mgc~benefits from searching locally and achieves a larger test statistic  than \Dcorr's, which can be very small in these nonlinear relationships. For type $20$, the test statistics for both \Mgc~and \Dcorr~are almost $0$ as expected. On the other hand, \Pearson's test statistic is large whenever there exists certain linear association, and almost $0$ otherwise. The comparison of sample statistics indicate that \Dcorr~may have inferior finite-sample testing power in nonlinear relationships, but a strong dependency signal is actually hidden in a local structure that \Mgc~may recover.  

\begin{figure}
\includegraphics[width=1.0\textwidth,trim={0cm 0cm 0cm 0cm},clip]{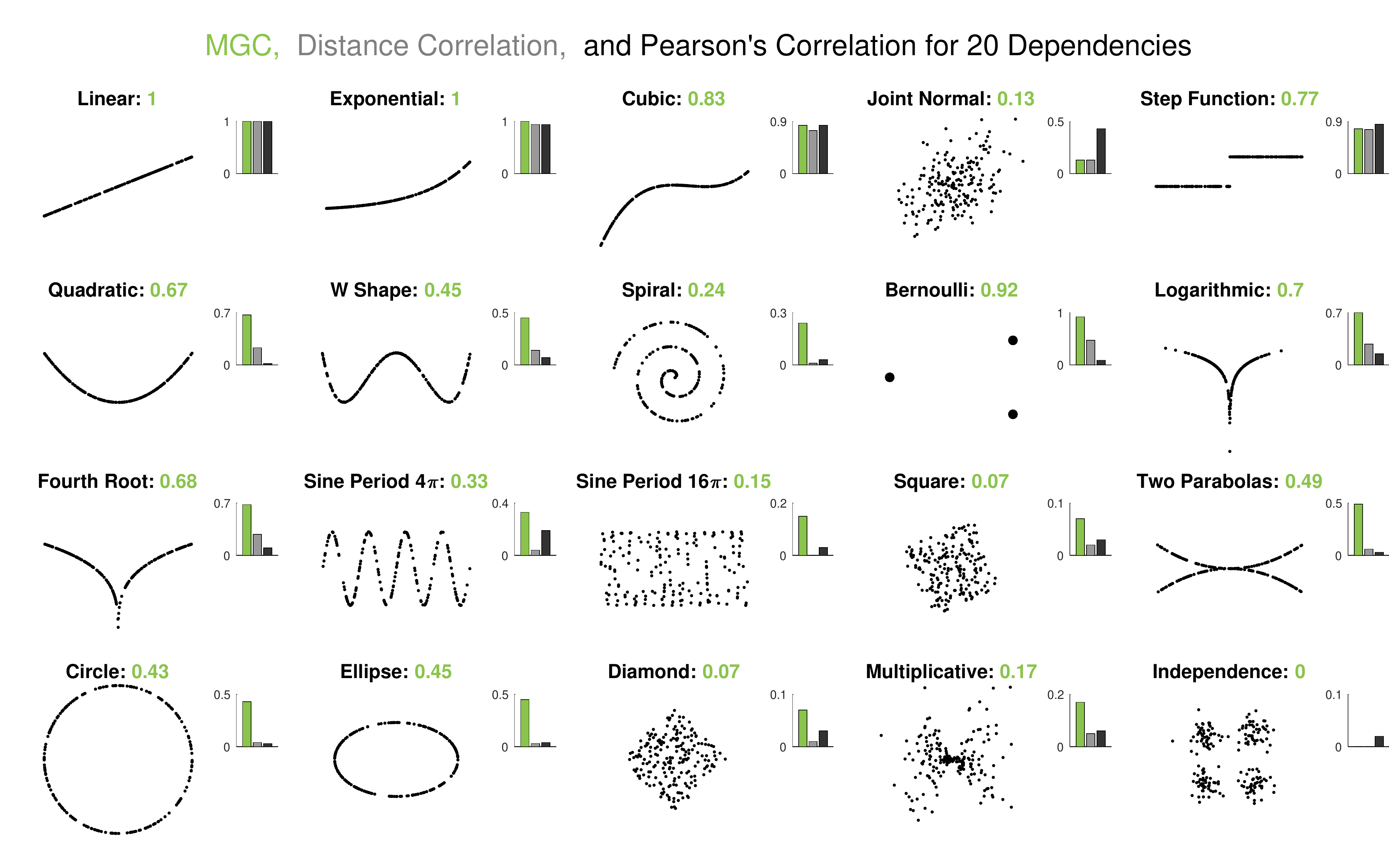}
\caption{For each panel, a pair of dependent $(\mathcal{X}_{n}, \mathcal{Y}_{n})$ at $n=100$ and $p=q=1$ is generated and visualized; the accompanying color bar compares \Mgc~(green), \Dcorr~(gray), and \Pearson~in the absolute value (black), all of which lie in the range of $[0,1]$ with $0$ indicating no relationship.  \Mgc~yields a non-zero sample correlation for each dependency, while being almost $0$ under independence. In comparison, the distance correlation can be close to $0$ for common nonlinear dependencies, while the Pearson's correlation only measures linear association and cannot capture nonlinear dependencies. The Sample \Mgc~statistic is shown above each panel.}
\label{f:dependencies}
\end{figure} 

\subsection*{Finite-Sample Testing Power}
\begin{figure}[!ht]
\centering
\includegraphics[width=0.48\textwidth,trim={1.5cm 0 0cm 0cm},clip]{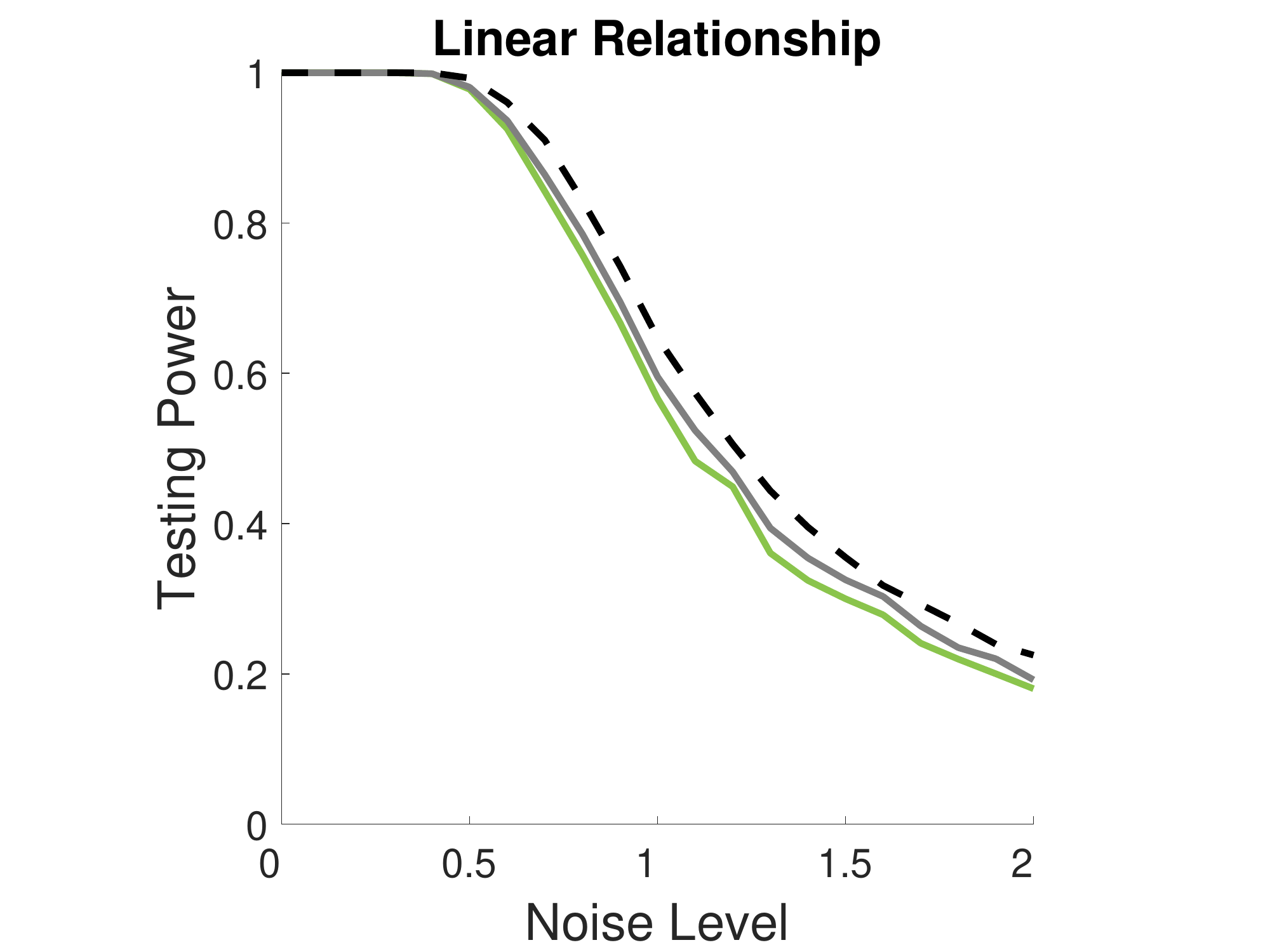}
\includegraphics[width=0.48\textwidth,trim={1.5cm 0 0cm 0cm},clip]{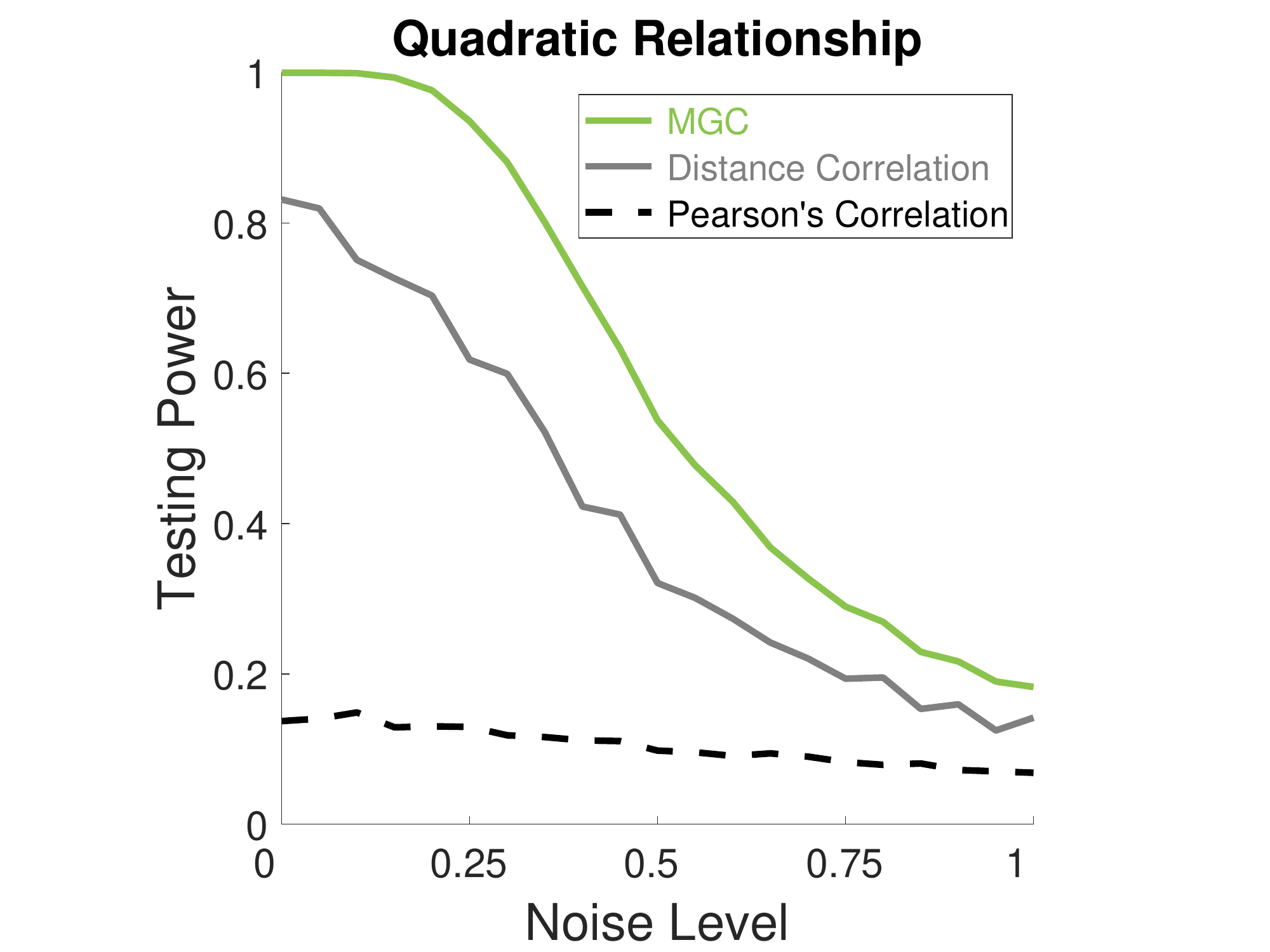}
  \caption{Comparing the power of \Mgc, \Dcorr, and \Pearson~in noisy linear relationship (left), and noisy quadratic relationship (right). For the linear relationship at $n=20$ and $p=q=1$, all three methods are almost the same with \Pearson~being slightly higher power; for the quadratic relationship, \Mgc~has a much higher power than \Dcorr~and \Pearson. The phenomenon is consistent throughout the remaining dependent simulations: for testing in monotonic relationships, \Pearson, \Dcorr, and \Mgc~almost coincide with each other; for strongly nonlinear relationships, \Mgc~almost always supersedes \Dcorr, and \Dcorr~is better than \Pearson.
}
\label{f:noise}
\end{figure}

Figure~\ref{f:noise}  shows the finite-sample testing power of \Mgc, \Dcorr, and \Pearson~for a linear  and a quadratic relationship at $n=20$ and $p=q=1$ with white noise (controlled by a constant). The testing power of \Mgc~is estimated as follows: we first generate dependent sample data $(\mathcal{X}_{n},\mathcal{Y}_{n})$ for $r=10,000$ replicates, compute Sample \Mgc~for each replicate to estimate the alternative distribution of \Mgc. Then we generate independent sample data $(\mathcal{X}_{n},\mathcal{Y}_{n})$ using the same marginal distributions for $r=10,000$ replicates, compute Sample \Mgc~to estimate the null distribution, and estimate the testing power at type $1$ error level $\alpha=0.05$. The testing power of \Dcorr~is estimated in the same manner, while the testing power of \Pearson~is directly computed via the t-test. \Mgc~has the best power in the quadratic relationship, while being almost identical to \Dcorr~and \Pearson~in the linear relationship.

\begin{figure}[htbp]
\includegraphics[width=1.0\textwidth,trim={0 0.5cm 3.2cm 0},clip]{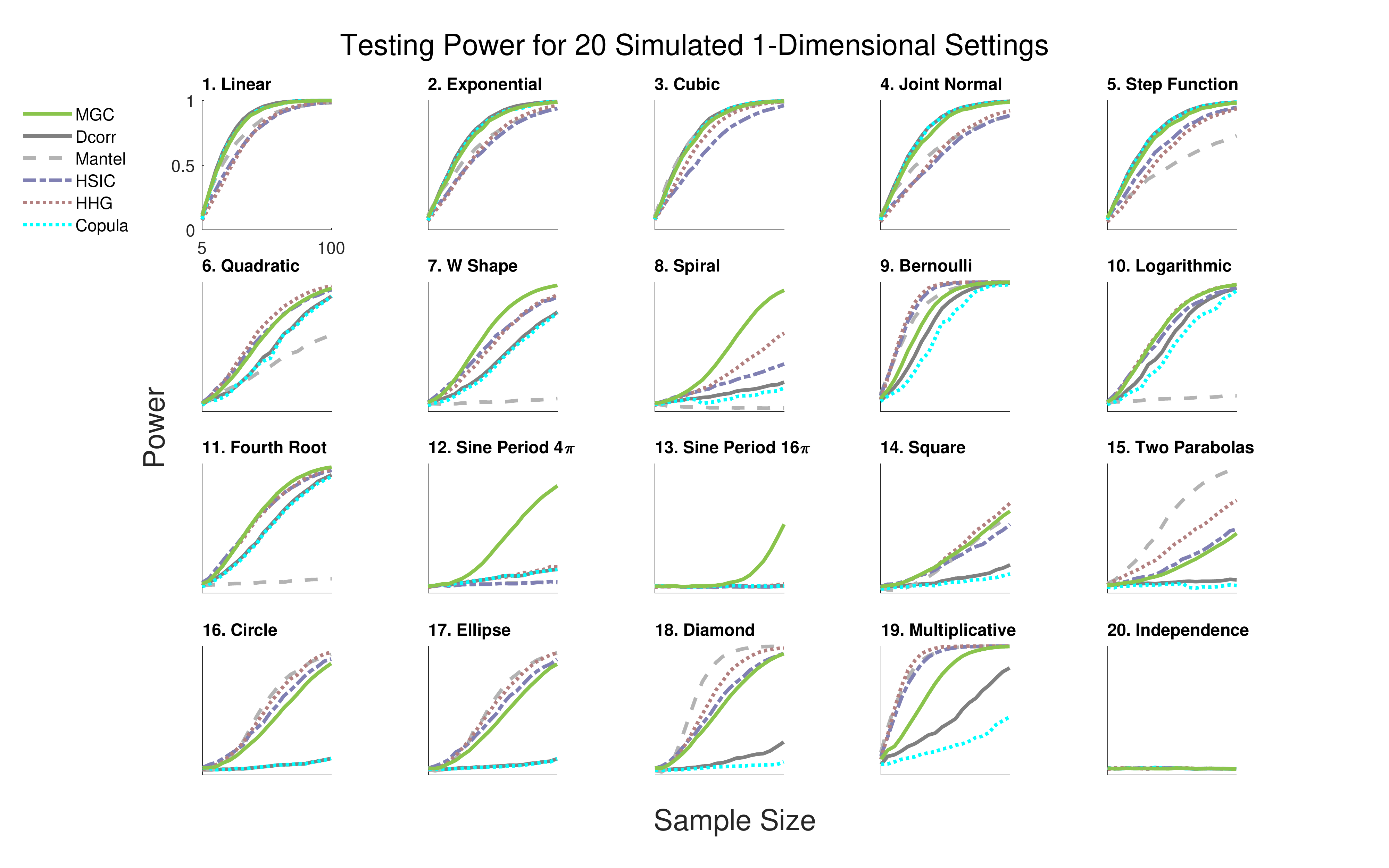}
\caption{
Comparing the testing power of \Mgc, \Dcorr, \Mantel, \Hsic, \Hhg, and \Copula.
for $20$ different univariate simulations.
Estimated via $10,000$ replicates of repeatedly generated dependent and independent sample data, each panel shows the estimated testing power at the type $1$ error level $\alpha=0.05$ versus sample sizes ranging from $n=5$ to  $100$. Excluding the independent simulation (\#20) where all methods yield power $0.05$, \Mgc~exhibits the highest or nearly highest power in most dependencies. 
Note that we only show the ticks for the first panel, because they are the same for every panel, i.e., the x-axis always ranges from $5$ to $100$ while the y-axis always ranges from $0$ to $1$.}
\label{f:1DAll}
\end{figure}

The same phenomenon holds throughout all the simulations we considered, i.e., \Mgc~achieves  almost the same power as \Dcorr~in monotonic relationships, while being able to improve the power in monotonic and strongly nonlinear relationships. The testing power of \Mgc~versus all other methods are shown in Figure~\ref{f:1DAll} for the univariate settings, and we plot the power versus the sample size from $5$ to $100$ for each simulation. 
Note that the noise level is tuned for each dependency for illustration purposes.

\begin{figure}[htbp]
\includegraphics[width=1.0\textwidth,trim={0 0.5cm 3.2cm 0},clip]{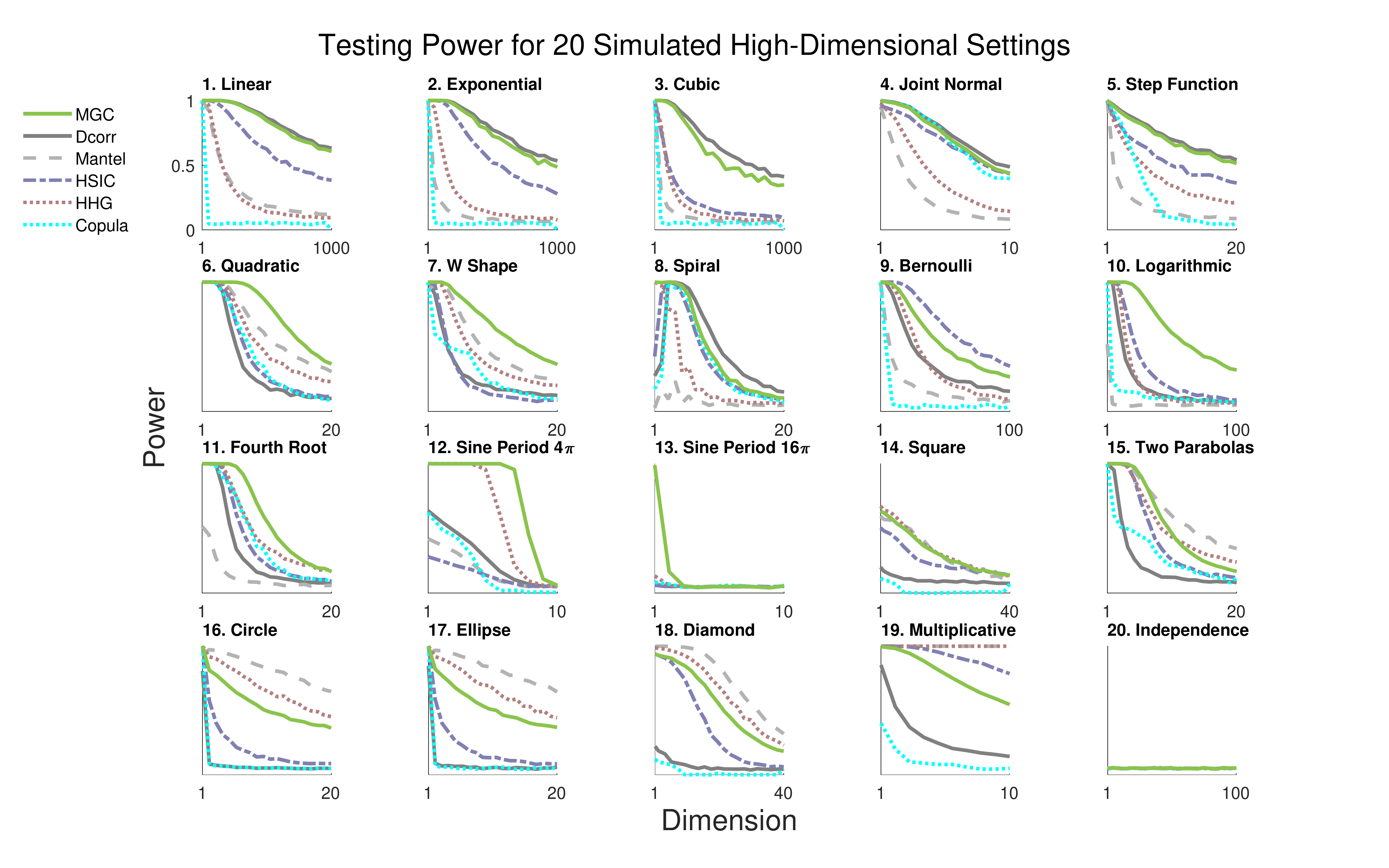}
\caption{The testing power computed in the same procedure as in Figure~\ref{f:1DAll}, except the $20$ simulations are now run at fixed sample size $n=100$ and increasing dimensionality $p$. Again, \Mgc~empirically achieves similar or higher power than the previous popular approaches for all dimensions on most settings. 
The ticks for y axis is only shown in the first panel, as the power has the same range in $[0,1]$ for every panel.}
\label{f:nDAll}
\end{figure}

\begin{figure}[!ht]
\centering
\includegraphics[width=0.48\textwidth,trim={1.5cm 0 0cm 0cm},clip]{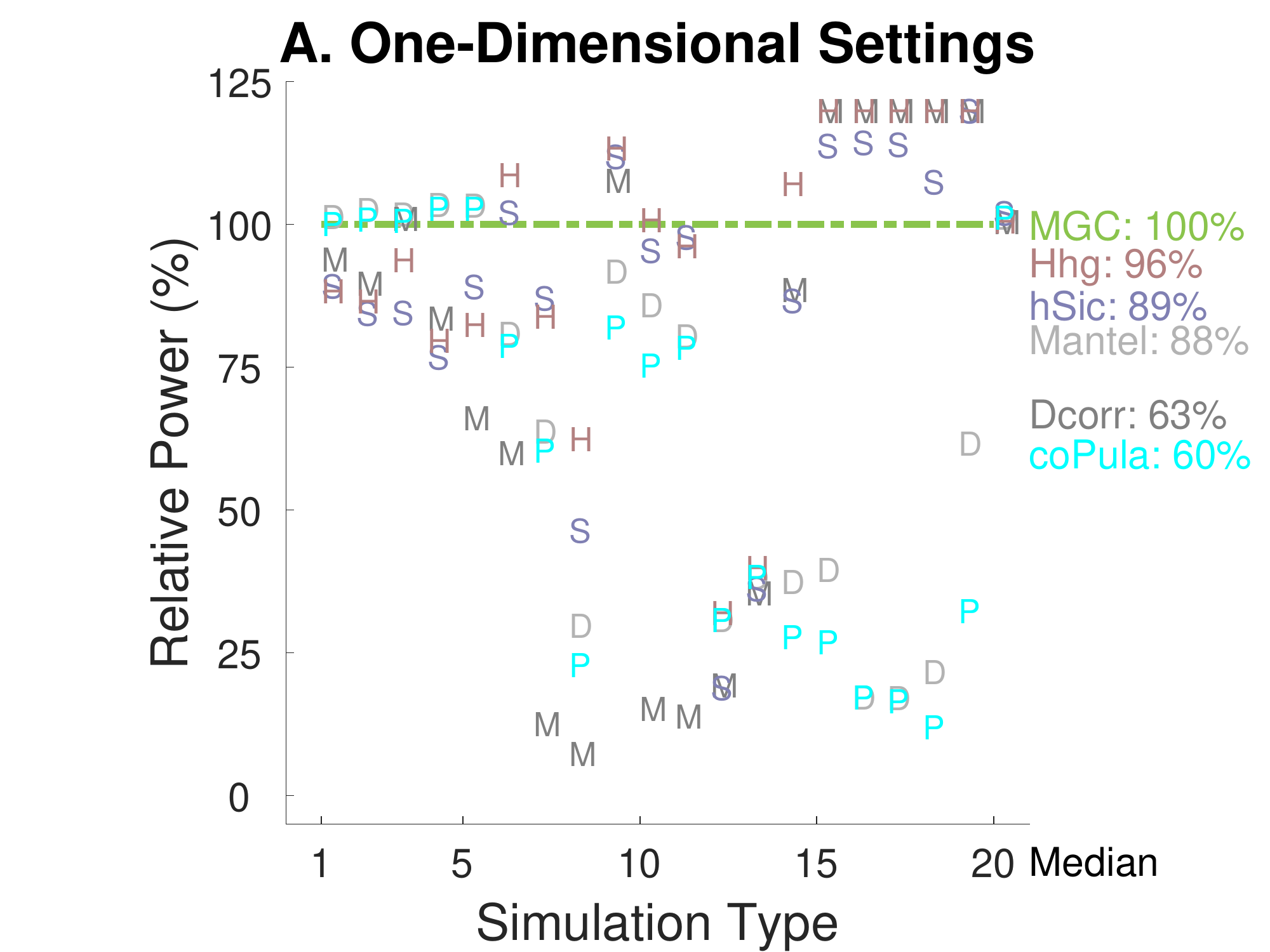}
\includegraphics[width=0.48\textwidth,trim={1.5cm 0 0cm 0cm},clip]{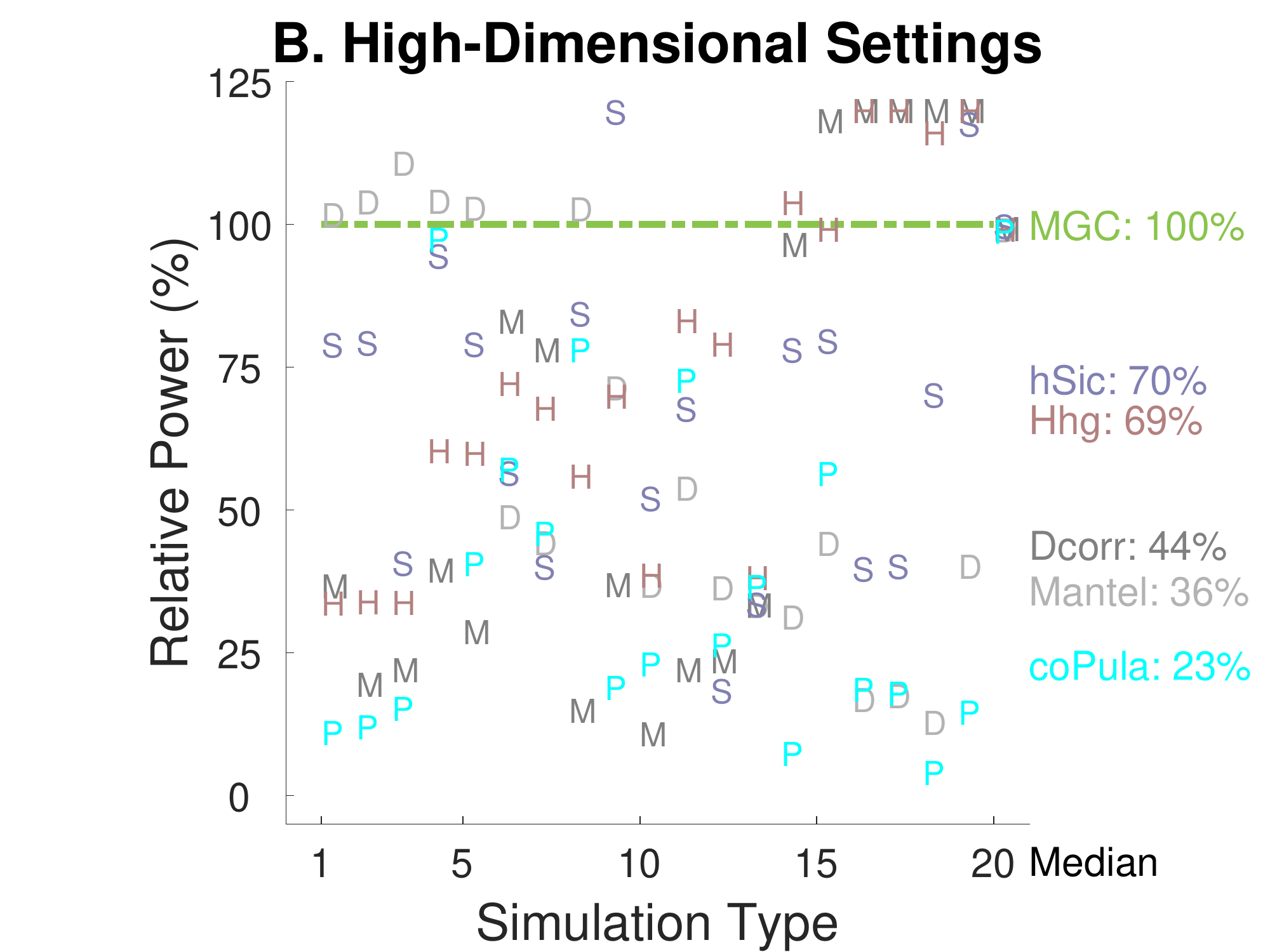}
  \caption{The relative Power of \Mgc~to other methods for testing  the $20$ simulations under one-dimensional and high-dimensional scenarios. (Left) For each simulation type, we average the testing power of each method in Figure~\ref{f:1DAll} over the sample size, then divide each average power by the average power of \Mgc. The last column (which also serves as the legend) shows the median power among all relative powers of type $1-19$. The same for the right panel, except it averages over the dimensionality in Figure~\ref{f:nDAll}. The relative power percentage indicates that \Mgc~is a very powerful method for finite-sample testing.
}
\label{f:Summary2}
\end{figure}

Figure~\ref{f:nDAll} compares the testing performance for the same $20$ simulations 
with a fixed sample size $n=100$
 and increasing dimensionality. The relative powers in the univariate and multivariate settings are then summarized in Figure~\ref{f:Summary2}. \Mgc~is overall the most powerful method, followed by \Hhg~and \Hsic. Since non-monotone relationships are prevalent among the $20$ settings, it is not a surprise that \Dcorr~is overall worse than \Hhg~and \Hsic, both of which also excel at nonlinear relationships.

Note that the same $20$ simulations were also used in \cite{ShenEtAl2016} for evaluation purposes. The main difference is that the Sample \Mgc~algorithm is now based on the improved threshold with theoretical guarantee. Comparing to the previous algorithm, the new threshold slightly improves the testing power in monotonic relationships (the first $5$ simulations).

\subsection*{Running Time}
\label{sec:time}

Sample \Mgc~can be computed and tested in the same running time complexity as distance correlation: Assume $p$ is the maximum feature dimension of the two datasets, distance computation and centering takes $\mc{O}(n^2 p)$, the ranking process takes $\mc{O}(n^2 \log n)$, all local covariances and correlations can be incrementally computed in $O(n^2)$ (the pseudo-code is shown in \cite{ShenEtAl2016}), the thresholding step of Sample \Mgc~takes $O(n^2)$ as well. Overall, Sample \Mgc~can be computed in $\mc{O}(n^2 \max\{\log n,p\})$. In comparison, the \Hhg~statistic requires the same complexity as \Mgc, while distance correlation saves on the $\log n$ term. 

\begin{figure}
\centering
\includegraphics[width=0.48\textwidth,trim={0cm 0 0cm 0cm},clip]{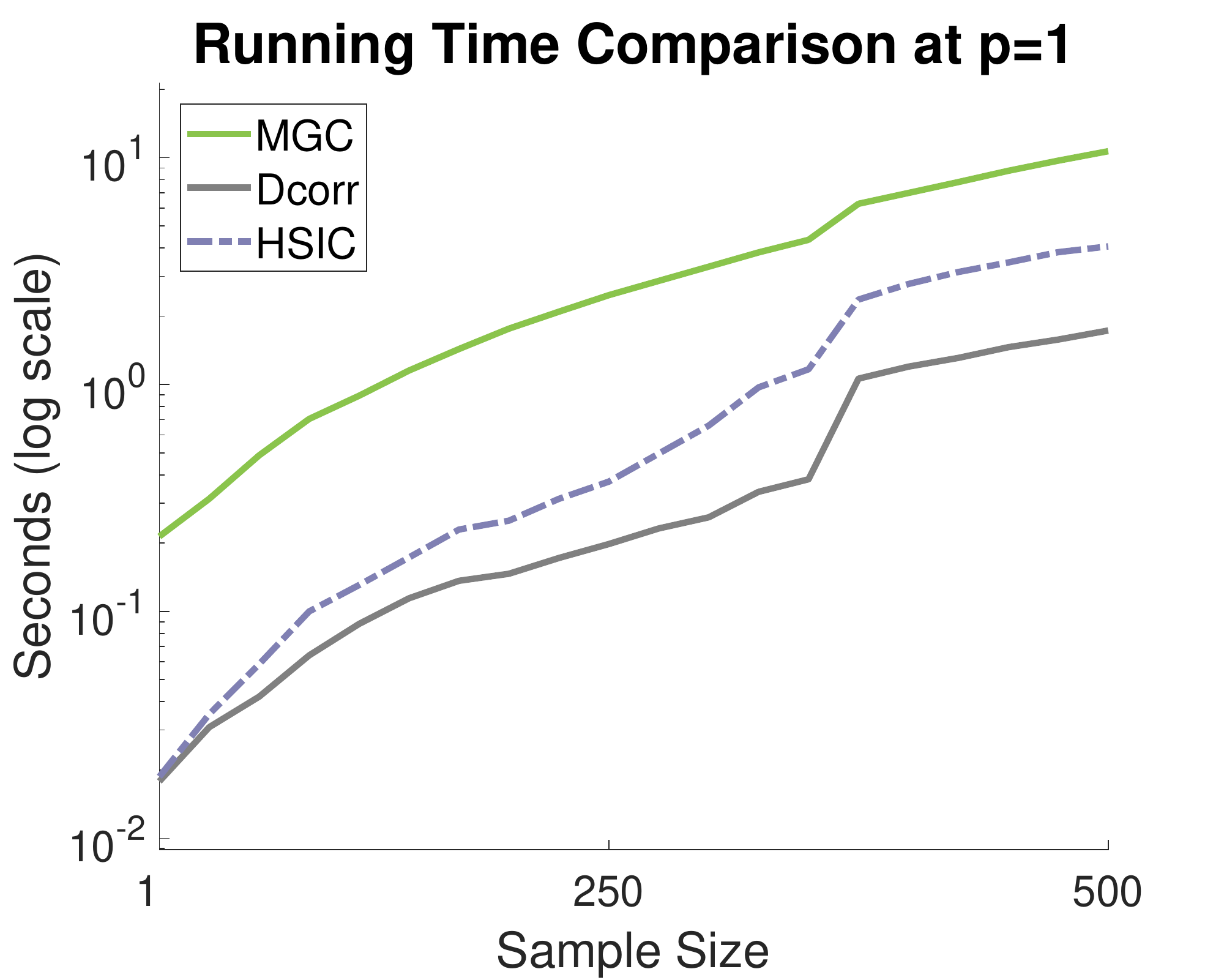}
\includegraphics[width=0.48\textwidth,trim={0cm 0 0cm 0cm},clip]{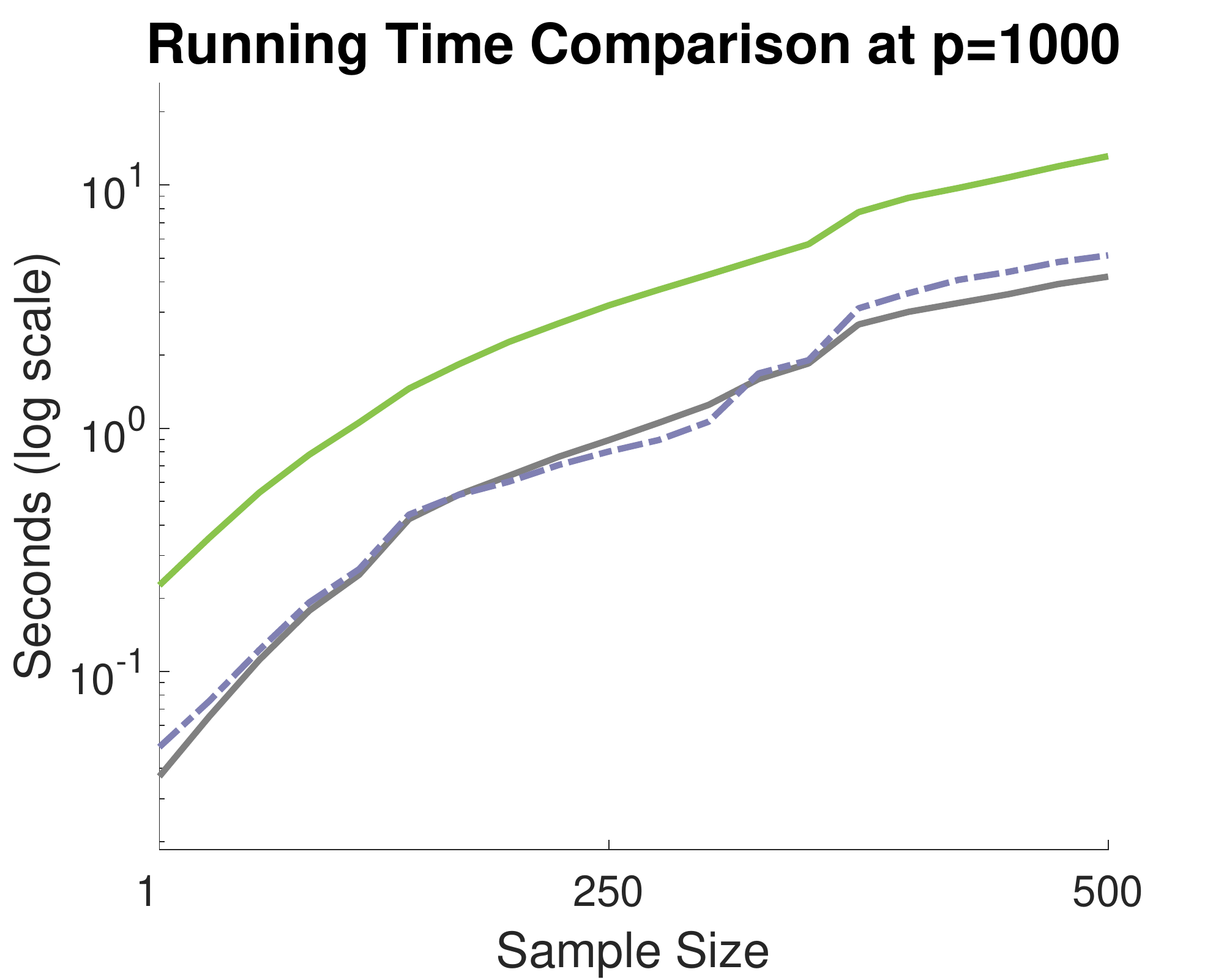}
\caption{Compute the test statistics of \Mgc, \Dcorr, and \Hsic~for $100$ replicates, then plot the average running time in log scale (clocked using Matlab 2017a on a Windows 10 machine with I7 six-core CPU). The sample data is repeatedly generated using the quadratic relationship in Appendix~\ref{appen:function}, the sample size increases from $25$ to $500$, and the dimensionality is fixed at $p=1$ on the left and $p=1000$ on the right. In either panel, the three lines differ by some constants in the log scale, suggesting the same running time complexity but different constants. \Mgc~has a higher intercept than the other two, which translates to about a constant of $6$ times of \Dcorr~and $3$ times of \Hsic~at $n=500$ and $p=1$, and about $3$ at $p=1000$. Note that the increase in $p$ has a relatively small effect in the running time, because the dimensionality $p$ takes part only in the distance matrix computation and is thus relatively cheap.}
\label{f:time}
\end{figure}

As the only part of \Mgc~that has the additional $\log n$ term is the column-wise ranking process, a multi-core architecture can reduce the running time to $\mc{O}(n^2 \max\{\log n,p\}/T)$. By making $T=\log(n)$ ($T$ is no more than $30$ at $1$ billion samples), \Mgc~effectively runs in $\mc{O}(n^2 p)$ and is of the same complexity as \Dcorr. The permutation test multiplies another $r$ to all terms except the distance computation, so overall the \Mgc~testing procedure requires $\mc{O}(n^2 \max\{r,p\})$, which is the same as \Dcorr, \Hhg, and \Hsic. Figure~\ref{f:time} shows that \Mgc~has approximately the same complexity as \Dcorr, and is slower by a constant in the actual running time.

\section{Conclusion}
\label{sec:dis}
In this paper, we formalize the population version of local correlation and \Mgc, connect them to the sample counterparts, prove the convergence and almost unbiasedness from the sample version to the population version, as well as a number of desirable properties for a well-defined correlation measure. In particular, population \Mgc~equals $0$ and the sample version converges to $0$ if and only if independence, making Sample \Mgc~valid and consistent under the permutation test. Moreover, Sample \Mgc~is designed in a computationally efficient manner, and the new threshold choice achieves both theoretical and empirical improvements. The numerical experiments confirm the empirical advantages of \Mgc~in a wide range of linear, nonlinear, high-dimensional dependencies. 

There are many potential future avenues to pursue. Theoretically, proving when and how one method dominates another in testing power is highly desirable. As the methods in comparison have distinct formulations and different properties, it is often difficult to compare them directly. However, a relative efficiency analysis may be viable when limited to methods of similar properties, such as \Dcorr~and \Hsic, or local statistic and global statistic. In terms of the locality principle, the geometric meaning of the local scale in \Mgc~is intriguing --- for example, does the family of local correlations fully characterize the joint distribution, and what is the relationship between the optimal local scale and the dependency geometry --- answering these questions may lead to further improvement of \Mgc, and potentially make the family of local correlations a valuable tool beyond testing. 

Method-wise, there are a number of alternative implementations that may be pursued. For example, the sample local correlations can be defined via $\epsilon$ ball instead of nearest neighbor graphs, i.e., truncate large distances based on absolute magnitude instead of the nearest neighbor graph. The maximization and thresholding mechanism may be further improved, e.g., thresholding based on the covariance instead of correlation, or design a better regularization scheme. There are many alternative approaches that can maintain consistency in this framework, and it will be interesting to investigate a better algorithm. In particular, we name our method as ``multiscale graph correlation" because the local correlations are computed via the k-nearest neighbor graphs, which is one way to generalize the distance correlation.

Application-wise, the \Mgc~method can directly facilitate new discoveries in many kinds of scientific fields, especially data of limited sample size and high-dimensionality such as in neuroscience and omics \cite{ShenEtAl2016}. Within the domain of statistics and machine learning, \Mgc~can be a very competitive candidate in any methodology that requires a well-defined dependency measure, e.g., variable selection \cite{LiZhongZhu2012}, time series \cite{Zhou2012}, etc. Moreover, the very idea of locality may improve other types of distance-based tests, such as the energy distance for K-sample testing \cite{SzekelyRizzo2013b}.

\clearpage
\bibliographystyle{ieeetr}
\bibliography{MGCbib}

\if1\blind
{
\section*{Acknowledgment}
This work was partially supported by
%
the National Science Foundation award DMS-1712947,
%
%
%
and the Defense Advanced Research Projects Agency's (DARPA) SIMPLEX program through SPAWAR contract N66001-15-C-4041.
%
%
%
%
The authors are grateful to the anonymous reviewers for the invaluable feedback leading to significant improvement of the manuscript, and thank Dr. Minh Tang and Dr. Shangsi Wang for useful discussions and suggestions.
}\fi

\clearpage
\appendix
\setcounter{figure}{0}
\renewcommand{\thealgorithm}{C\arabic{algorithm}}
\renewcommand{\thefigure}{E\arabic{figure}}
\renewcommand{\thesubsection}{\thesection.\arabic{subsection}}
\renewcommand{\thesubsubsection}{\thesubsection.\arabic{subsubsection}}
\pagenumbering{arabic}
\renewcommand{\thepage}{\arabic{page}}

\bigskip
\begin{center}
{\large\bf APPENDIX}
\end{center}

\section{Proofs}
\label{sec:proofs}

\subsection*{Theorem~\ref{thm1}}
\begin{proof}
Equation~\ref{eq:dcov1} defines the local covariance as \begin{align*}
dCov^{\rho_k, \rho_l}(\mbx,\mby) = \int_{\mathbb{R}^{p}\times \mathbb{R}^{q}} E(h^{\rho_{k}}_{\mbx}(t) \overline{h^{\rho_{l}}_{\mby'}(s)})-E(h^{\rho_{k}}_{\mbx}(t))E(h^{\rho_{l}}_{\mby'}(s)) w(t, s)dtds.
\end{align*}
Expanding the first integral term yields
\begin{align*}
&\int E(h^{\rho_{k}}_{\mbx}(t) \overline{h^{\rho_{l}}_{\mby'}(s)}) w(t, s)dtds \\
=&\ E(\int (g_{\mbx}(t)\overline{g_{\mbx'}(t)}-g_{\mbx}(t)\overline{g_{\mbx''}(t)}) (\overline{g_{\mby'}(s)}g_{\mby}(s)-\overline{g_{\mby'}(s)}g_{\mby'''}(s)) w(t, s)dtds \cdot \mb{I}_{\mbx,\mbx'}^{\rho_{k}} \mb{I}_{\mby',\mby}^{\rho_{l}}) \\
=&\ E(\int g_{\mbx \mby}(t,s) \overline{g_{\mbx' \mby'}(t,s)} w(t, s)dtds \cdot \mb{I}_{\mbx,\mbx'}^{\rho_{k}} \mb{I}_{\mby',\mby}^{\rho_{l}}) \\
& -E( \int g_{\mbx \mby}(t,s) \overline{g_{\mbx''}(t)g_{\mby'}(s)} w(t, s)dtds \cdot \mb{I}_{\mbx,\mbx'}^{\rho_{k}} \mb{I}_{\mby',\mby}^{\rho_{l}}) \\
& - E( \int \overline{g_{\mbx' \mby'}(t,s)} g_{\mbx}(t)g_{\mby'''}(s) w(t, s)dtds \cdot \mb{I}_{\mbx,\mbx'}^{\rho_{k}} \mb{I}_{\mby',\mby}^{\rho_{l}})\\
& + E( \int g_{\mbx}(t)g_{\mby'''}(s) \overline{ g_{\mbx''}(t) g_{\mby'}(s) } w(t, s)dtds \cdot  \mb{I}_{\mbx,\mbx'}^{\rho_{k}} \mb{I}_{\mby',\mby}^{\rho_{l}}) \\
=&\ E( \| \mbx-\mbx' \| \| \mby-\mby' \| \mb{I}_{\mbx,\mbx'}^{\rho_{k}} \mb{I}_{\mby',\mby}^{\rho_{l}} )  - E( \| \mbx-\mbx'' \| \| \mby-\mby' \| \mb{I}_{\mbx,\mbx'}^{\rho_{k}} \mb{I}_{\mby',\mby}^{\rho_{l}} ) \\
& - E( \| \mbx'-\mbx \| \| \mby'-\mby''' \| \mb{I}_{\mbx,\mbx'}^{\rho_{k}} \mb{I}_{\mby',\mby}^{\rho_{l}} ) + E( \| \mbx-\mbx'' \| \mb{I}_{\mbx,\mbx'}^{\rho_{k}} \| \mby'-\mby''' \|  \mb{I}_{\mby',\mby}^{\rho_{l}} ) \\
=&\ E(d^{\rho_{k}}_{\mbx} d^{\rho_{l}}_{\mby'}).
\end{align*}
Every other step being routine, the third equality transforms the $w(t,s)$ integral to Euclidean distances via the same technique employed in Remark 1 and the proof of Theorem 8 in \cite{SzekelyRizzo2009}. Also note that all four expectations are finite. For example, the first expectation in the third equality is finite, because $\| \mbx-\mbx' \| \| \mby-\mby' \|$ is always non-negative, and $E( \| \mbx-\mbx' \| \| \mby-\mby' \|)$ is non-negative and finite by the finite second moments assumption on \mbx~and \mby, such that 
\begin{align*}
0 \leq E( \| \mbx-\mbx' \| \| \mby-\mby' \| \mb{I}_{\mbx,\mbx'}^{\rho_{k}} \mb{I}_{\mby',\mby}^{\rho_{l}} ) \leq E( \| \mbx-\mbx' \| \| \mby-\mby' \|),
\end{align*}
which can be similarly established for the other three expectations.

The second integral term can be decomposed into 
\begin{align*}
\int E(h^{\rho_{k}}_{\mbx}(t))E(h^{\rho_{l}}_{\mby'}(s)) w(t, s)dtds = \int E(h^{\rho_{k}}_{\mbx}(t))w(t, s)dtds \cdot \int E(h^{\rho_{l}}_{\mby'}(s)) w(t, s)dtds,
\end{align*}
because the first expectation only has $t$ and the second expectation only has $s$, and $w(t,s)$ is a product of $t$ and $s$. Then 
\begin{align*}
&\int E(h^{\rho_{k}}_{\mbx}(t)) w(t, s)dtds = E(\int g_{\mbx}(t)\overline{g_{\mbx'}(t)}-g_{\mbx}(t)\overline{g_{\mbx''}(t)} w(t, s)dtds \cdot \mb{I}_{\mbx,\mbx'}^{\rho_{k}}) \\
=& E(\int g_{\mbx}(t)\overline{g_{\mbx'}(t)}  w(t, s)dtds \cdot \mb{I}_{\mbx,\mbx'}^{\rho_{k}}) - E(\int g_{\mbx}(t)\overline{g_{\mbx''}(t)}  w(t, s)dtds \cdot \mb{I}_{\mbx,\mbx'}^{\rho_{k}}) \\
=& E( \| \mbx-\mbx' \| \mb{I}_{\mbx,\mbx'}^{\rho_{k}})  - E( \| \mbx-\mbx'' \| \mb{I}_{\mbx,\mbx'}^{\rho_{k}}) \\
=& E(d^{\rho_{k}}_{\mbx}),
\end{align*}
where the two expectations involved are also finite. Similarly $\int E(\overline{h^{\rho_{l}}_{\mby'}(s)}) w(t, s)dtds = E( \| \mby'-\mby \| \mb{I}_{\mby',\mby}^{\rho_{l}})  - E( \| \mby'-\mby''' \| \mb{I}_{\mby',\mby}^{\rho_{l}})=E(d^{\rho_{l}}_{\mby'})$. Thus 
\begin{align*}
\int E(h^{\rho_{k}}_{\mbx}(t))E(h^{\rho_{l}}_{\mby'}(s)) w(t, s)dtds &= E(d^{\rho_{k}}_{\mbx})E(d^{\rho_{l}}_{\mby'}).
\end{align*}

Combining the results verifies that Equation~\ref{eq:dcov2} equals Equation~\ref{eq:dcov1}. Moreover, as every term in Equation~\ref{eq:dcov2} is of real-value, local covariance, variance, correlation are all real numbers.
\end{proof}

\subsection*{Theorem~\ref{thm2}}
\begin{proof}
When $\mbx$ and $\mby$ are independent, 
\begin{align*}
\int E(h^{\rho_{k}}_{\mbx}(t)\overline{h^{\rho_{l}}_{\mby'}(s)})w(t, s)dtds =\int E(h^{\rho_{k}}_{\mbx}(t))E(\overline{h^{\rho_{l}}_{\mby'}(s)})w(t, s)dtds, 
\end{align*}
thus $dCov^{\rho_{k}, \rho_{l}}(\mbx,\mby)=0$ at any $(\rho_k,\rho_l)$. So is the local correlation at any $(\rho_k,\rho_l) \in \mathcal{S}_{\epsilon}$.

To show the local covariance at the maximal scale $(\rho_k,\rho_l)=(1,1)$ equals the distance covariance, we
proceed via the alternative definition in Theorem~\ref{thm1}:
\begin{align*}
&dCov^{\rho_{k}=1, \rho_{l}=1}(\mbx,\mby) =E(d^{\rho_{k}}_{\mbx} d^{\rho_{l}}_{\mby'}) \\
=&\ E( \| \mbx-\mbx' \| \| \mby-\mby' \| )  - E( \| \mbx-\mbx'' \| \| \mby-\mby' \|) \\
&- E( \| \mbx'-\mbx \| \| \mby'-\mby''' \| ) + E( \| \mbx-\mbx'' \| ) E(\| \mby'-\mby''' \|  ) \\
=&\ E( \| \mbx-\mbx' \| \| \mby-\mby' \| )  - E( \| \mbx-\mbx'' \| \| \mby-\mby' \|) \\
&- E( \| \mbx-\mbx' \| \| \mby-\mby'' \| ) + E( \| \mbx-\mbx'' \| ) E(\| \mby-\mby'' \|  ) \\
=&\ dCov(\mbx,\mby),
\end{align*}
where the first equality follows by noting that $E(d^{\rho_{k}}_{\mbx})=E(d^{\rho_{l}}_{\mby'})=0$ at $\rho_{k}=\rho_{l}=1$, the second equality holds by switching the random variable notations within each expectation, and the last equality is the alternative definition of distance covariance in Theorem 8 of \cite{SzekelyRizzo2009}. It follows that $dVar^{\rho_{k}=1}(\mbx)=dVar(\mbx)$, $dVar^{\rho_{l}=1}(\mby)=dVar(\mby)$, and $dCorr^{\rho_{k}=1, \rho_{l}=1}(\mbx,\mby)=dCorr(\mbx,\mby)$. 

\end{proof}

\subsection*{Theorem~\ref{thmMax}}
\begin{proof}
Given two continuous random variables $(\mbx,\mby)$, we first illustrate the continuity of local covariance with respect to $\rho_{k}$ at fixed $\rho_{l}$: For any $\delta$ with the understanding that $\rho_{k} \pm \delta \in [0,1]$, we have
\begin{align*}
dCov^{\rho_k+\delta, \rho_l}(\mbx,\mby) -dCov^{\rho_k, \rho_l}(\mbx,\mby) = E((d^{\rho_{k}+\delta}_{\mbx}-d^{\rho_{k}}_{\mbx}) d^{\rho_{l}}_{\mby'}) - E(d^{\rho_{k}+\delta}_{\mbx}-d^{\rho_{k}}_{\mbx}) E(d^{\rho_{l}}_{\mby'}),
\end{align*}
where the expectation is taken with respect to all random variables inside, and
\begin{align*}
d^{\rho_{k}+\delta}_{\mbx}&=(\|\mbx-\mbx'\|-\|\mbx-\mbx''\|) \mb{I}_{\mbx,\mbx'}^{\rho_{k}+\delta}\\
d^{\rho_{k}}_{\mbx}&=(\|\mbx-\mbx'\|-\|\mbx-\mbx''\|) \mb{I}_{\mbx,\mbx'}^{\rho_{k}}
\end{align*}
Then Cauchy-Schwarz and finite second moment of $\mbx$ yield that
\begin{align*}
&\lim_{\delta \rightarrow 0} |E(d^{\rho_{k}+\delta}_{\mbx}-d^{\rho_{k}}_{\mbx})|^2 \\
\leq &\ E\{(\|\mbx-\mbx'\|-\|\mbx-\mbx''\|)^2\} \lim_{\delta \rightarrow 0} E(|\mb{I}_{\mbx,\mbx'}^{\rho_{k}+\delta}-\mb{I}_{\mbx,\mbx'}^{\rho_{k}}|^2)\\
=&\ 0.
\end{align*}
Moreover, the finite second moment of \mby~guarantees finiteness of $E( d^{\rho_{l}}_{\mby'})$ and
\begin{align*}
&\lim_{\delta \rightarrow 0} |E((d^{\rho_{k}+\delta}_{\mbx}-d^{\rho_{k}}_{\mbx}) d^{\rho_{l}}_{\mby'})|^2 \\
\leq &\ E\{(\|\mbx-\mbx'\|-\|\mbx-\mbx''\|)^2 {d^{\rho_{l}}_{\mby'}}^2\} \lim_{\delta \rightarrow 0} E(|\mb{I}_{\mbx,\mbx'}^{\rho_{k}+\delta}-\mb{I}_{\mbx,\mbx'}^{\rho_{k}}|^2)\\
=&\ 0,
\end{align*}
which leads to the continuity of local covariance with respect to $\rho_{k}$:
\begin{align*}
\lim_{\delta \rightarrow 0} dCov^{\rho_k+\delta, \rho_l}(\mbx,\mby) - dCov^{\rho_k, \rho_l}(\mbx,\mby)=0.
\end{align*}

The same holds for fixed $\rho_{k}$ such that 
\begin{align*}
\lim_{\delta \rightarrow 0} dCov^{\rho_k, \rho_l+\delta}(\mbx,\mby) - dCov^{\rho_k, \rho_l}(\mbx,\mby)=0.
\end{align*}
Applying the above yields that
\begin{align*}
&dCov^{\rho_k+\delta_{1}, \rho_l+\delta_{2}}(\mbx,\mby) -dCov^{\rho_k, \rho_l}(\mbx,\mby)\\
= &\ dCov^{\rho_k+\delta_{1}, \rho_l+\delta_{2}}(\mbx,\mby) -dCov^{\rho_k, \rho_l+\delta_{2}}(\mbx,\mby) + dCov^{\rho_k, \rho_l+\delta_{2}}(\mbx,\mby) - dCov^{\rho_k, \rho_l}(\mbx,\mby)\\
\rightarrow & \  0 \mbox { for any $\delta_{1}$ and $\delta_{2}$ satisfying $|\delta_{1}+\delta_{2}| \rightarrow 0$.}
\end{align*}
So the local covariance is continuous with respect to $(\rho_{k},\rho_{l}) \in [0,1]\times[0,1]$. The continuity of the local variance can be shown similarly, and it follows that the local correlation is continuous in $\mathcal{S}_{\epsilon}$.

At $\rho_{k}=1$, $dVar^{\rho_k}(\mbx)=dVar(\mbx) \geq 0$ with equality if and only if $\mbx$ is a constant, and $\mathcal{S}_{\epsilon}$ is empty in the trivial case. Otherwise by the continuity of local variance, for any $\epsilon < dVar(\mbx)$ there exists $\epsilon_{k}$ such that for all $\rho_{k} \in [\epsilon_{k},1]$, $dVar^{\rho_k}(\mbx) \geq \epsilon$. Same for $dVar^{\rho_l}(\mby)$, thus $\mathcal{S}_{\epsilon}$ is non-empty except when either random variable is a constant. It follows that the local correlation is continuous within the non-empty and compact domain $\mathcal{S}_{\epsilon}$, and extreme value theorem ensures the existence of population \Mgc~and the optimal scale.
\end{proof}


\subsection*{Theorem~\ref{thm3}}
\begin{proof}
By Theorem~\ref{thm2} and definition of \Mgc, it holds that
\begin{align*}
\GG^{*}(\mbx,\mby) \geq dCorr^{\rho_{k}=\rho_{l}=1}(\mbx,\mby)=dCorr(\mbx,\mby).
\end{align*}
When $\mbx$ and $\mby$ are independent, all local correlations are $0$ by Theorem~\ref{thm2}, so $\GG^{*}(\mbx,\mby)=0$ as well. When dependent, distance correlation is larger than $0$, and it follows that $\GG^{*}(\mbx,\mby) \geq dCorr(\mbx,\mby)>0$. Therefore, \Mgc~equals $0$ if and only if independence, just like the distance correlation.
\end{proof}

\subsection*{Theorem~\ref{thm4}}
\begin{proof}
We prove this theorem by three steps: \textbf{(i)}, the expectation of the sample local covariance is shown to equal the population local covariance; \textbf{(ii)}, the variance of the sample statistic is of $\mathcal{O}(\frac{1}{n})$; \textbf{(iii)}, sample local covariance is shown to convergence to the population counterpart uniformly. Then the convergence trivially extends to the sample local variance and correlation.

\textbf{(i)}: Expanding the first and second term of population local covariance in Equation~\ref{eq:dcov2}, we have $E(d^{\rho_{k}}_{\mbx} d^{\rho_{l}}_{\mby'})=\alpha_{1}-\alpha_{2}-\alpha_{3}+\alpha_{4}$ with
\begin{align*}
\alpha_{1}&=E( \| \mbx-\mbx' \| \| \mby-\mby' \| \mb{I}_{\mbx,\mbx'}^{\rho_{k}} \mb{I}_{\mby',\mby}^{\rho_{l}} ),\\
\alpha_{2}&=E( \| \mbx-\mbx'' \| \| \mby-\mby' \| \mb{I}_{\mbx,\mbx'}^{\rho_{k}} \mb{I}_{\mby',\mby}^{\rho_{l}} ),\\
\alpha_{3}&=E( \| \mbx'-\mbx \| \| \mby'-\mby''' \| \mb{I}_{\mbx,\mbx'}^{\rho_{k}} \mb{I}_{\mby',\mby}^{\rho_{l}} ),\\
\alpha_{4}&=E( \| \mbx-\mbx'' \| \| \mby'-\mby''' \| \mb{I}_{\mbx,\mbx'}^{\rho_{k}} \mb{I}_{\mby',\mby}^{\rho_{l}}),
\end{align*}
and $E(d^{\rho_{k}}_{\mbx})E(d^{\rho_{l}}_{\mby'})=\alpha_{5}-\alpha_{6}-\alpha_{7}+\alpha_{8}$ with
\begin{align*}
\alpha_{5}&=E( \| \mbx-\mbx' \| \mb{I}_{\mbx,\mbx'}^{\rho_{k}}) E(\| \mby-\mby' \|  \mb{I}_{\mby',\mby}^{\rho_{l}}),\\
\alpha_{6}&=E( \| \mbx-\mbx' \| \mb{I}_{\mbx,\mbx'}^{\rho_{k}}) E(\| \mby''-\mby' \|  \mb{I}_{\mby',\mby}^{\rho_{l}}),\\
\alpha_{7}&=E( \| \mbx-\mbx'' \| \mb{I}_{\mbx,\mbx'}^{\rho_{k}}) E(\| \mby-\mby' \|  \mb{I}_{\mby',\mby}^{\rho_{l}}),\\
\alpha_{8}&=E( \| \mbx-\mbx'' \| \mb{I}_{\mbx,\mbx'}^{\rho_{k}}) E(\| \mby''-\mby' \|  \mb{I}_{\mby',\mby}^{\rho_{l}}).
\end{align*}
All the $\alpha$'s are bounded due to the finite first moment assumption on $(\mbx,\mby)$. Note that for distance covariance, one can go through the same proof with only three terms -- $\alpha_{1}, \alpha_{2}, \alpha_{5}$ -- while the local version involves eight terms, due to the additional random variables for local scales.

For the sample local covariance, the expectation of the first term can be expanded as
\begin{align*}
&\ \frac{1}{n(n-1)}\sum_{i \neq j}^{n}E(A_{ij}B_{ji}\mb{I}(R^{A}_{ij} \leq k)\mb{I}(R^{B}_{ji} \leq l))\\
=&\ E( (\frac{n-2}{n-1}\tilde{A}_{ij}-\frac{1}{n-1}\sum_{s \neq i,j} \tilde{A}_{sj}) \\
&\ \cdot (\frac{n-2}{n-1}\tilde{B}_{ji}-\frac{1}{n-1}\sum_{s \neq i,j} \tilde{B}_{si}) \mb{I}(R^{A}_{ij} \leq k)\mb{I}(R^{B}_{ji} \leq l))\\
=&\ \frac{(n-2)^2}{(n-1)^2} (\alpha_{1}-\alpha_{2}-\alpha_{3})+\frac{(n-2)(n-3)}{(n-1)^2}\alpha_{4}+\mathcal{O}(\frac{1}{n}) \\
=&\ \alpha_{1}-\alpha_{2}-\alpha_{3}+\alpha_{4}+\mathcal{O}(\frac{1}{n}).
\end{align*}
The expectation of the second term can be similarly expanded as
\begin{align*}
& E(\frac{1}{n(n-1)}\sum_{i \neq j}^{n}A^{k}_{ij} \frac{1}{n(n-1)}\sum_{i \neq j}^{n}B^{l}_{ji}) \\
=&\ \frac{1}{n^2(n-1)^2} \sum_{u\neq v}^{n} E(A_{uv}\mb{I}(R^{A}_{uv} \leq k) \sum_{i \neq j}^{n}B_{ji}\mb{I}(R^{B}_{ji} \leq l))\\
=&\ \frac{1}{n(n-1)} E( (\frac{n-2}{n-1}\tilde{A}_{uv}-\frac{1}{n-1}\sum_{s \neq u,v} \tilde{A}_{sv})\mb{I}(R^{A}_{uv} \leq k)  \\
&\cdot \sum_{i \neq j}^{n} (\frac{n-2}{n-1}\tilde{B}_{ji}-\frac{1}{n-1}\sum_{s \neq i,j} \tilde{B}_{si}) \mb{I}(R^{B}_{ji} \leq l)\\
=&\ \alpha_{5}-\alpha_{6}-\alpha_{7}+\alpha_{8}+\mathcal{O}(\frac{1}{n}).
\end{align*}
Combining the results yields that $E(dCov^{k,l}(\mathcal{X}_{n},\mathcal{Y}_{n}))=dCov^{\rho_{k},\rho_{l}}(\mbx,\mby)+\mathcal{O}(\frac{1}{n})$.

\textbf{(ii)}: The variance of sample local covariance is computed as
\begin{align*}
&Var( \E( A^{k}- \E(A^{k})) (B^{l'}- \E(B^{l'})))\\
=&\ \frac{1}{n^2(n-1)^2} Var(\sum_{i \neq j}^{n} (A^{k}_{ij}- \E(A^{k})) (B^{l}_{ji}- \E(B^{l})))\\
=&\ \frac{n^4}{n^2(n-1)^2} \mathcal{O}(\frac{1}{n}) + \frac{n^3}{n^2(n-1)^2} \mathcal{O}(1).
\end{align*}
The last equality follows because: there are $n^4$ covariance terms in the numerator of $\mathcal{O}(\frac{1}{n})$, because $Cov( (A^{k}_{ij}- \E(A^{k})) (B^{l}_{ji}- \E(B^{l})), (A^{k}_{uv}- \E(A^{k})) (B^{l}_{vu}- \E(B^{l})))$ are only related via the column centering when $(i,j)$ does not equal $(u,v)$; and there remains $n^3$ covariance terms of at most $\mathcal{O}(1)$. Note that the finite second moment assumption of $(\mbx,\mby)$ is required for the big $\mathcal{O}$ notation to have a bounding constant. Therefore, the variance of sample local covariance is of $\mathcal{O}(\frac{1}{n})$.

\textbf{(iii)}: $dCov^{k,l}(\mathcal{X}_{n},\mathcal{Y}_{n})$ converges to the population local covariance by applying the strong law of large numbers on U-statistics \cite{KoroljukBook}. Namely, the first term of sample local covariance satisfies
\begin{align*}
&\frac{1}{n(n-1)}\sum_{i \neq j}^{n}A_{ij}B_{ji}\mb{I}(R^{A}_{ij} \leq k)\mb{I}(R^{B}_{ji} \leq l) \\
=&\ \frac{1}{n}\sum_{i=1}^{n}(\frac{1}{n-1}\sum_{j \neq i}^{n} (\frac{n-2}{n-1}\tilde{A}_{ij}-\frac{1}{n-1}\sum_{s \neq i,j} \tilde{A}_{sj}) \\
& \cdot (\frac{n-2}{n-1}\tilde{B}_{ji}-\frac{1}{n-1}\sum_{s \neq i,j} \tilde{B}_{si}) \mb{I}(R^{A}_{ij} \leq k)\mb{I}(R^{B}_{ji} \leq l) ) \\
\rightarrow &\ \frac{1}{n}\sum_{i=1}^{n} (\alpha_{1|(x_i,y_i)}-\alpha_{2|(x_i,y_i)}-\alpha_{3|(x_i,y_i)}+\alpha_{4|(x_i,y_i)})\\
\rightarrow &\ \alpha_{1}-\alpha_{2}-\alpha_{3}+\alpha_{4},
\end{align*} 
where the second line applies law of large numbers at each $i$ by conditioning on $(\mbx,\mby)=(x_i,y_i)$ for each $\alpha$'s, and the last line follows by applying law of large numbers to the independently distributed conditioned $\alpha$'s. Similarly, the second term of sample local covariance can be shown to converge to the second term in population local covariance. The convergence is also uniform: each local covariance are dependent with each other, and actually repeats the summands with each other. Thus there exists a scale $(k,l)$ such that $dCor^{k,l}$ has the largest deviation from the mean than all other local covariances, and one can find a suitable $\epsilon$ to bound the maximum deviation for all $dCor^{k,l}$.

Alternatively, convergence in probability can be directly established from (i) and (ii) by applying the Chebyshev's inequality; the almost sure convergence can also be proved via the integral definition using almost the same steps as in Theorems 1 and 2 from \cite{SzekelyRizzoBakirov2007}, i.e., first define the empirical characteristic function via the $w$ integral for the sample local covariance, and show it converges to the population local covariance in Equation~\ref{eq:dcov1} by the law of large numbers on U-statistics.
\end{proof}

\subsection*{Corollary~\ref{thm5}}
\begin{proof}
It follows directly from Theorem~\ref{thm2}, Theorem~\ref{thm4}, and the convergence of sample distance correlation to the population \cite{SzekelyRizzoBakirov2007}. 
\end{proof}

\subsection*{Corollary~\ref{cor1}}
\begin{proof}
The population \Mantel~and its equivalence to expectation of Euclidean distances can be established via the same steps as in Theorem~\ref{thm1}. The convergence of sample \Mantel~to its population version can be derived based on either the same procedure in Theorem~\ref{thm4}, or Theorems 1 and 2 from \cite{SzekelyRizzoBakirov2007} with minimal notational changes.
\end{proof}

\subsection*{Theorem~\ref{thm6}}
\begin{proof}

\textbf{(a)}: Regardless of the threshold choice, the algorithm enforces Sample \Mgc~to be always no less than $dCorr^{n,n}(\mathcal{X}_{n},\mathcal{Y}_{n})$, and no more than $\max\{dCorr^{k,l}(\mathcal{X}_{n},\mathcal{Y}_{n})\}$. 

\textbf{(b)}: By Corollary~\ref{thm5}, $dCorr^{n,n}(\mathcal{X}_{n},\mathcal{Y}_{n}) \rightarrow dCorr(\mbx,\mby)$, then the uniform convergence by Theorem~\ref{thm4} ensures that $\max\{dCorr^{k,l}(\mathcal{X}_{n},\mathcal{Y}_{n})\} \rightarrow \GG^{*}(\mbx,\mby)$. When $\mbx$ and $\mby$ are independent, $dCorr(\mbx,\mby)$ and $\GG^{*}(\mbx,\mby)$ are both $0$, to which Sample \Mgc~must converge; when dependent, $dCorr^{n,n}(\mathcal{X}_{n},\mathcal{Y}_{n})$ converges to a positive constant, so Sample \Mgc~must converge to a constant that is either the same or larger.
\end{proof}

\subsection*{Theorem~\ref{thm7}}
\begin{proof}
\textbf{(a)}: Given $\GG^{*}(\mbx,\mby)>dCorr(\mbx,\mby)$, by the continuity of local correlations with respect to $(\rho_{k},\rho_{l})$, there always exists a non-empty connected area $\mathcal{R} \in \mathcal{S}_{\epsilon}$ such that $dCorr^{\rho_k,\rho_l}(\mbx,\mby)>dCorr(\mbx,\mby)$ for all $(\rho_{k},\rho_{l}) \in \mathcal{R}$. Among all possible areas we take the one with largest area.

As $n$ increases to infinity, the set $\{(\frac{k-1}{n-1},\frac{l-1}{n-1}) \ | \ (k, l) \in [n]^2\}$ is a dense subset of $[0,1] \times [0,1]$, and $\{dCorr^{k,l}(\mathcal{X}_{n},\mathcal{Y}_{n})\}$ is also a dense subset of $\{dCorr^{\rho_{k},\rho_{l}}(\mbx,\mby)\}$. Thus for $n$ sufficiently large, the area $\mathcal{R}$ can always be approximated via the largest connected component $R$ by the Sample \Mgc~algorithm. As all sample local correlations within the region $R$ are larger than the sample distance correlation, so is the smoothed maximum. Note that if the threshold $\tau_n$ does not converge to $0$, e.g., if $\tau_n$ is a positive constant like $0.05$, Sample \Mgc~will fail to identify a region $R$ when $0.05 > \GG^{*}(\mbx,\mby)$. 

\textbf{(b)}: Following (a), if optimal scale of \Mgc~is in the largest area $\mathcal{R}$, the sample maximum within $R$ converges to the true maximum within $\mathcal{R}$, i.e., Sample \Mgc~converges to the population \Mgc.
\end{proof}

\subsection*{Corollary~\ref{cor2}}
\begin{proof}
For $v=\frac{n(n-3)}{2}$, $z \sim Beta(\frac{v-1}{2})$, the convergence of $\tau_n= 2F^{-1}_{z}(1-\frac{0.02}{n}) -1$ can be shown as follows: by computing the variance of the Beta distribution and using Chebyschev's inequality, it follows that
\begin{align*}
& \frac{0.04}{n} = Prob(|z-0.5| \geq \tau_n /2) \leq \mathcal{O}(\frac{1}{n^2 \tau_{n}^{2}})\\
\Rightarrow &\ \tau_{n}=\mathcal{O}(\frac{1}{\sqrt{n}}) \rightarrow 0.
\end{align*} 
The equation also implies that the percentile choice can be either fixed or anything no larger than $1-\frac{c}{n^2}$ for some constant $c$, beyond which the convergence of $\tau_n$ to $0$ will be broken.
\end{proof}

\subsection*{Theorem~\ref{thm8}}
\begin{proof}
To prove consistency under the permutation test, it suffices to show that at any type $1$ error level $\alpha$, the p-value of \Mgc~is asymptotically less than $\alpha$. The p-value can be expressed by:
\begin{align*}
& Prob(\GG^{*}(\mathcal{X}_{n},\mathcal{Y}_{n}^{\pi}) > \GG^{*}(\mathcal{X}_{n},\mathcal{Y}_{n})) \\
= &\ \sum_{j=0}^{n} Prob(\GG^{*}(\mathcal{X}_{n},\mathcal{Y}_{n}^{\pi}) > \GG^{*}(\mathcal{X}_{n},\mathcal{Y}_{n}) | \pi \mbox{ is a partial derangement of size $j$})\\
& \times Prob(\mbox{partial derangement of size $j$})  
\end{align*}
by conditioning on the permutation being a partial derangement of size $j$, e.g., $j=0$ means $\pi$ is a derangement, while $j=n$ means $\pi$ does not permute any position. 

As $n \rightarrow \infty$, we always have
\begin{align*}
&Prob(\mbox{partial derangement of size $j$}) \rightarrow e^{-1} / j!, \\
&\GG^{*}(\mathcal{X}_{n},\mathcal{Y}_{n}) \rightarrow \epsilon >0 \ \mbox{ under dependence.}
\end{align*}
Thus it suffices to show that for any $\epsilon >0$,
\begin{align}
\label{eq:permconsistency}
\lim_{n\rightarrow \infty} e^{-1}\sum_{j=0}^{n} Prob(\GG^{*}(\mathcal{X}_{n},\mathcal{Y}_{n}^{\pi}) > \epsilon | \mbox{ partial derangement of size $j$}) / j! \rightarrow 0.
\end{align}
Then we decompose the above summations into two different cases. The first case is when $j$ is a fixed size, $\mathcal{X}_{n}$ and $\mathcal{Y}_{n}^{\pi}$ are asymptotically independent (due to the \emph{iid} assumption), thus $\GG^{*}(\mathcal{X}_{n},\mathcal{Y}_{n}^{\pi})$ converges to $0$. The other case is the remaining partial derangements $\pi$ of size $\mathcal{O}(n)$, but these partial derangements occur with probability converging to $0$, i.e., for any $\alpha > 0$, there exists $N_{1}$ such that 
\begin{align*}
e^{-1} \sum_{j=N_{1}+1}^{+\infty} 1/j! < \alpha / 2,
\end{align*}
as $\sum\limits_{j=0}^{n} 1/j!$ is bounded above and converges to $e$. Then back to the first case, there further exists $N_{2}>N_{1}$ such that for any $j\leq N_{1}$ and all $n > N_{2}$
\begin{align*}
Prob(\GG^{*}(\mathcal{X}_{n},\mathcal{Y}_{n}^{\pi}) > \epsilon)| \mbox{ partial derangement of size $j$}) < \alpha / 2.
\end{align*}
It follows that for all $n > N_{2}$,
\begin{align*}
& e^{-1} \sum_{j=0}^{n} Prob(\GG^{*}(\mathcal{X}_{n},\mathcal{Y}_{n}^{\pi}) > \epsilon | \mbox{ partial derangement of size $j$}) / j!\\
< &\ e^{-1} \sum_{j=0}^{N_{1}} \alpha / 2 j! + e^{-1} \sum_{j=N_{1}+1}^{n} 1 / j!\\
< &\ \alpha.
\end{align*}
Thus the convergence in Equation~\ref{eq:permconsistency} holds.


Therefore, at any type $1$ error level $\alpha>0$, the p-value of Sample \Mgc~under the permutation test will eventually be less than $\alpha$ as $n$ increases, such that Sample \Mgc~always successfully detects any dependency. Thus Sample \Mgc~is consistent against all dependencies with finite second moments. 

When $\mbx$ and $\mby$ are independent, each column of $\mathcal{X}_{n}$ and the corresponding column of $\mathcal{Y}_{n}$ are independent for any permutation. Therefore, $\GG^{*}(\mathcal{X}_{n},\mathcal{Y}_{n}^{\pi})$ distributes the same as $\GG^{*}(\mathcal{X}_{n},\mathcal{Y}_{n})$ for any random permutation $\pi$, and $Prob(\GG^{*}(\mathcal{X}_{n},\mathcal{Y}_{n}^{\pi}) > \GG^{*}(\mathcal{X}_{n},\mathcal{Y}_{n}))$ is uniformly distributed in $[0,1]$. Thus Sample \Mgc~is valid.
\end{proof}

\subsection*{Lemma~\ref{lem1}}
\begin{proof}
\begin{align*}
dCov^{k,l}(\mathcal{X}_{n},\mathcal{Y}_{n}) &= \E(A^{k} \circ B^{l'})- \E(A^{k} \circ J)\E(B^{l} \circ J) \\
&= tr (A^{k}B^{l})- tr (A^{k}J)tr(B^{l}J)\\
&=tr [ (A^{k}- tr (A^{k}J)J) (B^{l}-tr(B^{l}J)J)]\\
& = \sum_{i=1}^{n} \lambda_{i} [ (A^{k}- tr (A^{k}J)J) (B^{l}-tr(B^{l}J)J)],
\end{align*}
where the first line is the definition, the second line follows by noting that $\E(A \circ B^{'})=tr(AB)$ and $\E(A)=\E(A \circ J)=tr(AJ)$ for any two matrices $A$ and $B$, and the last two lines follow from basic properties of matrix trace.
\end{proof}

\subsection*{Theorem~\ref{thm:dvar}}
\begin{proof}
For all these properties, it suffices to prove them on the sample local variance $dVar^{k}(\mathcal{X}_{n})$ first. Then the population version follows by the convergence property in Theorem~\ref{thm4}.

\textbf{(a)}: Based on Lemma~\ref{lem1} it holds that
\begin{align*}
dVar^{k}(\mathcal{X}_{n}) 
& = \sum_{i=1}^{n} \lambda^{2}_{i}[A^{k}- tr (A^{k}J)J] \geq 0.
\end{align*}

\textbf{(b)}: Following part (a), we have
\begin{align*}
&\ dVar^{k}(\mathcal{X}_{n})=0\\
\Leftrightarrow & \ \lambda_{i}[A^{k}- tr (A^{k}J)J] =0, \ \forall i \\
\Leftrightarrow & \ A^{k}- tr (A^{k}J)J = 0_{n \times n} \\
\Leftrightarrow & \ A^{k}_{ij} = tr (A^{k}J), \ \forall i,j =1,\ldots, n\\
\Leftrightarrow & \ A^{k}_{ij}=tr (A^{k}J)=0, \ \forall i,j =1,\ldots, n,
\end{align*}
where the last line follows by observing that $A^{k}_{ii}=0$ by Equation~\ref{localCoef2}. Therefore, distance variance equals $0$ if and only if $A^{k}$ is the zero matrix.

A trivial case is $k=0$, which corresponds to $\rho_k =0$ asymptotically. Otherwise $A^{k}$ is a zero matrix if and only if for all $(i,j)$ satisfying $\mb{I}(R^{A}_{ij} \leq k)=1$, 
\begin{align*}
\tilde{A}_{ij}=\frac{1}{n-1}\sum_{s=1}^{n} \tilde{A}_{sj}. 
\end{align*}
Namely, for each point $x_j$, its $k$ smallest distance entries all equal the mean distances with respect to $x_j$, which can only happen when $\tilde{A}_{ij}$ is a constant for all $i \neq j$ at a fixed $j$. Due to the symmetry of the distance matrix, all the off-diagonal entries of $\tilde{A}$ are the same, i.e., $\tilde{A}=u (J - I)$ for some constant $u \geq 0$.

When $u=0$, all observations are the same, so $\mbx$ is a constant. Otherwise all observations are equally distanced from each other by a distance of $u>0$, which occurs with probability $0$ under the \emph{iid} assumption. This is because when $\mbx^{'}$ and $\mbx^{''}$ are independent, one cannot have $\|\mbx^{''}-\mbx\|=\|\mbx^{'}-\mbx\|$ almost surely unless they are degenerate. 

From another point of view, for given sample data that happens to be equally distanced, e.g., $n$ points in $n-1$ dimensions, sample variances can still be $0$. But this scenario occurs with probability $0$ when each observation is assumed \emph{iid}.

\textbf{(c)}: This follows trivially from the definition, because upon the transformation the distance matrix is unchanged up-to a factor of $u$. 
\end{proof}

\subsection*{Theorem~\ref{thm:dcor}}
\begin{proof}
Similar as in Theorem~\ref{thm:dvar}, it suffices to prove (a) and (b) for the sample local correlation, then they automatically hold for the population version by convergence. 

\textbf{(a)}: The symmetric part is trivial: for any $(\rho_k,\rho_l) \in [0,1] \times [0,1]$, by Lemma~\ref{lem1}
\begin{align*}
dCov^{k,l}(\mathcal{X}_{n},\mathcal{Y}_{n}) 
&=tr [ (A^{k}- tr (A^{k}J)J) (B^{l}-tr(B^{l}J)J)] \\
&=tr [ (B^{l}-tr(B^{l}J)J)(A^{k}- tr (A^{k}J)J)] \\
&=dCov^{l,k}(\mathcal{Y}_{n},\mathcal{X}_{n}).
\end{align*}
Then by the Cauchy-Schwarz inequality on the trace,
\begin{align*}
|dCov^{k,l}(\mathcal{X}_{n},\mathcal{Y}_{n}) |
=& \ |tr [ (A^{k}- tr (A^{k}J)J) (B^{l}-tr(B^{l}J)J)]|\\
= & \ \sqrt{tr [ (A^{k}- tr (A^{k}J)J) (B^{l}-tr(B^{l}J)J)]}\\
& \ \times  \sqrt{tr [ (A^{k}- tr (A^{k}J)J) (B^{l}-tr(B^{l}J)J)]} \\
\leq & \ \sqrt{ dVar^{k}(\mathcal{X}_{n}) dVar^{l}(\mathcal{Y}_{n})}.
\end{align*}
Thus $dCorr^{k,l}(\mathcal{X}_{n},\mathcal{Y}_{n}) =dCorr^{l,k}(\mathcal{Y}_{n},\mathcal{X}_{n}) \in [-1,1]$.

\textbf{(b)}: 
The if direction is clear: under isometry, $\tilde{A}=|u| \tilde{B}$, both share the same k-nearest-neighbor graph, so that $A^{k} = |u| \cdot B^{k}$. Thus $dCov^{k,k}(\mathcal{X}_{n},\mathcal{Y}_{n}) = \frac{1}{u^2} dVar^{k}(\mathcal{X}_{n})= u^2 \cdot dVar^{k}(\mathcal{Y}_{n})$, and $dCorr^{k,k}(\mathcal{X}_{n},\mathcal{Y}_{n})=1$. For the only if direction: by part (a), the local correlation can be $\pm 1$ if and only if $(A^{k}- tr (A^{k}J)J)$ is a scalar multiple of $(B^{l}-tr(B^{l}J)J)$, say some constant $u$. 

First we argue that the non-zero entries in $A^{k}$ must match the non-zero entries in $B^{l}$. Namely, the k-nearest neighbor graph is the same between $\tilde{A}$ and $\tilde{B}$. As $A^{k}_{ii}=B^{l}_{ii}=0$, $-tr (A^{k}J)$ must be a scalar multiple of $-tr (B^{l}J)$. Then if there exists $i \neq j$ such that $A^{k}_{ij}=0$ while $B^{l}_{ij} \neq 0$, $-tr (A^{k}J)$ must be the same scalar multiple of $B^{l}_{ij}-tr (B^{l}J)$, which is not possible unless $B^{l}_{ij}=0$. Thus $k=l$ and $\mb{I}(R^{A}_{ij} \leq k)= \mb{I}(R^{B}_{ij} \leq k)$ for all $(i,j)$.

Next we show the scalar multiple must be positive, i.e., the local correlation cannot be $-1$. Assuming it can be $-1$, then 
\begin{align*}
& \ A^{k}- tr (A^{k}J)J=-|u|(B^{k}-tr(B^{k}J)J) \\
\Leftrightarrow & \ A^{k}+|u|B^{k}=(tr (A^{k}J)+|u|tr(B^{k}J))J \\
\Leftrightarrow & \ A^{k}+|u|B^{k}=0_{n \times n} \\
\Leftrightarrow & \ A+|u|B=0_{n \times n}, 
\end{align*}
where the second to last line follows because the diagonal entries of $A^{k}+|u|B^{k}$ are $0$ by definition, and the last line follows by observing that $tr (A^{k}J)$ and $tr (B^{k}J)$ are both negative unless $k=n$ (e.g., $A$ is always centered to have zero matrix mean, while $A^{k}$ keeps the $k$ smallest entries per column so its matrix mean is negative til $k=n$). However, if the last line is true, then the original distance correlation shall be $-1$, which cannot happen under the \emph{iid} assumption as shown in \cite{SzekelyRizzoBakirov2007}. Note that the derivation also shows that the local correlations can be $-1$ for general dissimilarity matrices without the \emph{iid} assumption, i.e., when $\tilde{A}+|u| \tilde{B}=v(J-I)$ for some constant $v$.

Therefore, the scalar multiple must be positive, and $A^{k}- tr (A^{k}J)J = |u| (B^{k}-tr (B^{k}J)J)$. As the diagonals satisfy $A^{k}_{ii}=B^{k}_{ii}=0$, it holds that $tr (A^{k}J)= |u| tr (B^{k}J)$ and $A^{k}=|u| B^{k}$. Thus for each $(i, j)$ satisfying $\mb{I}(R^{A}_{ij} \leq k)=1$:
\begin{align*}
& \ \tilde{A}_{ij}-\frac{1}{n-1}\sum_{s=1}^{n} \tilde{A}_{sj}=|u|(\tilde{B}_{ij}-\frac{1}{n-1}\sum_{s=1}^{n} \tilde{B}_{sj}) \\
\Leftrightarrow & \ \tilde{A}_{ij}-|u|\tilde{B}_{ij} = \frac{1}{n-1}\sum_{s=1}^{n} \tilde{A}_{sj} -\frac{|u|}{n-1}\sum_{s=1}^{n} \tilde{B}_{sj}\\
\Leftrightarrow & \ \tilde{A}_{ij}-|u|\tilde{B}_{ij} = v.
\end{align*}
We argue that if $\tilde{A}_{ij}=|u|\tilde{B}_{ij}+v$ for each $(i, j)$ satisfying $\mb{I}(R^{A}_{ij} \leq k)=1$, it also holds for all $(i,j)$. Suppose there exists $(s,j)$ with $\mb{I}(R^{A}_{sj} \leq k)=0$ and $\tilde{A}_{sj} = |u|\tilde{B}_{sj}+v+w$ for some $w \neq 0$. Without loss of generality, there must exist one more index $t$ such that $\mb{I}(R^{A}_{tj} \leq k)=0$ and $\tilde{A}_{tj} = |u|\tilde{B}_{tj}+v-w$ to maintain the mean (or multiple indices in a similar manner). This requires $\|\mbx^{''}-\mbx\|-|u|\|\mby^{''}-\mby\|=\|\mbx^{'}-\mbx\|-|u|\|\mby^{'}-\mby\|+2w$, so $(\mbx^{''},\mby^{''})$ and $(\mbx^{'},\mby^{'})$ are related by $w$ when conditioning on $(\mbx,\mby)$. Thus it imposes a dependency structure and violates the \emph{iid} assumption.

Therefore $\tilde{A}-|u|\tilde{B}=v(J-I)$. When $v=0$, $\tilde{A}=|u|\tilde{B}$ is equivalent to that $(\mbx, u \mby$) are related by an isometry. When $v \neq 0$, it requires each distance entries to be added by the same constant, which occurs with probability $0$ under the \emph{iid} assumption. Namely, if $\|\mbx^{'}-\mbx\|-|u| \|\mby^{'}-\mby\|=\|\mbx^{''}-\mbx\|-|u| \|\mby^{''}-\mby\|=v \neq 0$ almost surely, then $(\mbx^{''},\mby^{''})$ and $(\mbx^{'},\mby^{'})$ are related by $v$ when conditioning on $(\mbx,\mby)$, in which case these two pairs become dependent and the \emph{iid} assumption is violated. 

\textbf{(c)}: As each local correlation is symmetric and bounded for either population or sample case, \Mgc~is symmetric and within $[-1,1]$ by part (a).

\textbf{(d)}: If $\mbx$ and $u \mby$ are related by an isometry, the distance correlation (or the local correlation at the largest scale) equals $1$. For population, \Mgc~takes the maximum local correlation; for sample, \Mgc~cannot be smaller than the local correlation at the largest scale. In both cases population and Sample \Mgc~equal $1$.

When population or Sample \Mgc~equal $1$, there exists at least one local correlation that equals $1$, i.e., $dCorr^{k,l}(\mathcal{X}_{n},\mathcal{Y}_{n})=1$. From the inequality in part (a), $k$ must equal $l$ for the equality to hold. Otherwise the number of non-zero entries does not match between $A^{k}$ and $B^{l}$, and $A^{k}$ cannot be a scalar multiple of $B^{l}$. Thus there exists $k$ such that $dCorr^{k,k}(\mathcal{X}_{n},\mathcal{Y}_{n})=1$, and the conclusion follows from part (b).
\end{proof}

\section{Simulation Dependence Functions}
\label{appen:function}
This section presents the $20$ simulations used in the experiment section, which is mostly based on a combination of simulations from previous works \cite{SzekelyRizzoBakirov2007, SimonTibshirani2012, GorfineHellerHeller2012}. We only made changes to add noise and a weight vector for higher dimensions, thereby making them more difficult and easier to compare all methods throughout different dimensions and sample sizes. 

For the random variable $\mbx \in \Real^{p}$, we denote $\mbx_{[d]}, d=1,\ldots,p$ as the $d^{th}$ dimension of \mbx. For the purpose of high-dimensional simulations, $w \in \Real^{p}$ is a decaying vector with $w_{[d]}=1/d$ for each $d$, such that $w\TT \mbx$ is a weighted summation of all dimensions of \mbx.
Furthermore, $\mc{U}(a,b)$ denotes the uniform distribution on the interval $(a,b)$, $\mc{B}(p)$ denotes the Bernoulli distribution with probability $p$, $\mc{N}(\mu,{\Sigma})$ denotes the normal distribution with mean ${\mu}$ and covariance ${\Sigma}$,
$U$ and $V$ represent some auxiliary random variables, $\kappa$ is a scalar constant to control the noise level (which equals $1$ for one-dimensional simulations and $0$ otherwise), and $\epsilon$ is sampled from an independent standard normal distribution unless mentioned otherwise.


\setcounter{equation}{0}
\begin{compactenum}
\item Linear $(\mbx,\mby) \in \Real^{p} \times \Real$:
\begin{align*}
\mbx &\sim \mc{U}(-1,1)^{p},\\
\mby &=w\TT \mbx+\kappa\epsilon.
\end{align*}
\item Exponential $(\mbx,\mby) \in \Real^{p} \times \Real$:
\begin{align*}
\mbx &\sim \mc{U}(0,3)^{p}, \\
\mby &=exp(w\TT \mbx)+10\kappa\epsilon.
\end{align*}
\item Cubic $(\mbx,\mby) \in \Real^{p} \times \Real$:
\begin{align*}
\mbx &\sim \mc{U}(-1,1)^{p}, \\
\mby &=128(w\TT \mbx-\tfrac{1}{3})^3+48(w\TT \mbx-\tfrac{1}{3})^2-12(w\TT \mbx-\tfrac{1}{3})+80\kappa\epsilon.
\end{align*}
\item Joint normal $(\mbx,\mby) \in \Real^{p} \times \Real^{p}$: Let $\rho=1/2p$, $I_{p}$ be the identity matrix of size $p \times p$, $J_{p}$ be the matrix of ones of size $p \times p$, and $\Sigma = \begin{bmatrix} I_{p}&\rho J_{p}\\ \rho J_{p}& (1+0.5\kappa) I_{p} \end{bmatrix}$. Then
\begin{align*}
(\mbx, \mby) &\sim \mc{N}(0, \Sigma).
\end{align*}
\item Step Function $(\mbx,\mby) \in \Real^{p} \times \Real$:
\begin{align*}
\mbx &\sim \mc{U}(-1,1)^{p},\\
\mby &=\mb{I}(w\TT \mbx>0)+\epsilon,
\end{align*}
where $\mb{I}$ is the indicator function, that is $\mb{I}(z)$ is unity whenever $z$ true, and zero otherwise.
\item Quadratic $(\mbx,\mby) \in \Real^{p} \times \Real$:
\begin{align*}
\mbx &\sim \mc{U}(-1,1)^{p},\\
\mby&=(w\TT \mbx)^2+0.5\kappa\epsilon.
\end{align*}
\item W Shape $(\mbx,\mby) \in \Real^{p} \times \Real$:  $U \sim \mc{U}(-1,1)^{p}$,
\begin{align*}
\mbx &\sim \mc{U}(-1,1)^{p},\\
\mby&=4\left[ \left( (w\TT \mbx)^2 - \tfrac{1}{2} \right)^2 + w\TT U/500 \right]+0.5\kappa\epsilon.
\end{align*}
\item Spiral $(\mbx,\mby) \in \Real^{p} \times \Real$: $U \sim \mc{U}(0,5)$, $\epsilon \sim \mc{N}(0, 1)$,
\begin{align*}
\mbx_{[d]}&=U \sin(\pi U)  \cos^{d}(\pi U) \mbox{ for $d=1,\ldots,p-1$},\\
\mbx_{[p]}&=U \cos^{p}(\pi U),\\
\mby&= U \sin(\pi U) +0.4 p\epsilon.
\end{align*}
\item Uncorrelated Bernoulli $(\mbx,\mby) \in \Real^{p} \times \Real$: $U \sim \mc{B}(0.5)$, $\epsilon_{1} \sim \mc{N}(0, I_{p})$, $\epsilon_{2} \sim \mc{N}(0, 1)$,
\begin{align*}
\mbx &\sim \mc{B}(0.5)^{p}+0.5\epsilon_{1},\\
\mby&=(2U-1)w\TT \mbx+0.5\epsilon_{2}.
\end{align*}
\item Logarithmic $(\mbx,\mby) \in \Real^{p} \times \Real^{p}$: $\epsilon \sim \mc{N}(0, I_{p})$
\begin{align*}
\mbx &\sim \mc{N}(0, I_{p}),\\
\mby_{[d]}&=2\log_{2}(|\mbx_{[d]}|)+3\kappa\epsilon_{[d]} \mbox{ for $d=1,\ldots,p$.}
\end{align*}
\item Fourth Root $(\mbx,\mby) \in \Real^{p} \times \Real$:
\begin{align*}
\mbx &\sim \mc{U}(-1,1)^{p},\\
\mby&=|w\TT \mbx|^\frac{1}{4}+\frac{\kappa}{4}\epsilon.
\end{align*}
\item Sine Period $4\pi$ $(\mbx,\mby) \in \Real^{p} \times \Real^{p}$: $U \sim \mc{U}(-1,1)$, $V \sim \mc{N}(0,1)^{p}$, $\theta=4\pi$,
\begin{align*}
\mbx_{[d]}&=U+0.02 p V_{[d]} \mbox{ for $d=1,\ldots,p$}, \\
\mby&=\sin ( \theta \mbx )+\kappa\epsilon.
\end{align*}
\item Sine Period $16\pi$ $(\mbx,\mby) \in \Real^{p} \times \Real^{p}$: Same as above except $\theta=16\pi$ and the noise on $\mby$ is changed to $0.5\kappa\epsilon$.
\item Square $(\mbx,\mby) \in \Real^{p} \times \Real^{p}$: Let $U \sim \mc{U}(-1,1)$, $V \sim \mc{U}(-1,1)$, $\epsilon \sim \mc{N}(0,1)^{p}$, $\theta=-\frac{\pi}{8}$. Then
\begin{align*}
\mbx_{[d]}&=U \cos\theta + V \sin\theta + 0.05 p\epsilon_{[d]},\\
\mby_{[d]}&=-U \sin\theta + V \cos\theta,
\end{align*}
for $d=1,\ldots,p$.
\item Two Parabolas $(\mbx,\mby) \in \Real^{p} \times \Real$: $\epsilon \sim \mc{U}(0,1)$, $U \sim \mc{B}(0.5)$,
\begin{align*}
\mbx &\sim \mc{U}(-1,1)^{p},\\
\mby&=\left( (w\TT \mbx)^2  + 2\kappa\epsilon\right) \cdot (U-\tfrac{1}{2}).
\end{align*}
\item Circle $(\mbx,\mby) \in \Real^{p} \times \Real$: $U \sim \mc{U}(-1,1)^{p}$, $\epsilon \sim \mc{N}(0, I_{p})$, $r=1$,
\begin{align*}
\mbx_{[d]}&=r \left(\sin(\pi U_{[d+1]})  \prod_{j=1}^{d} \cos(\pi U_{[j]})+0.4 \epsilon_{[d]}\right) \mbox{ for $d=1,\ldots,p-1$},\\
\mbx_{[p]}&=r \left(\prod_{j=1}^{p} \cos(\pi U_{[j]})+0.4 \epsilon_{[p]}\right),\\
\mby&= \sin(\pi U_{[1]}).
\end{align*}
\item Ellipse $(\mbx,\mby) \in \Real^{p} \times \Real$: Same as above except $r=5$.
\item Diamond $(\mbx,\mby) \in \Real^{p} \times \Real^{p}$: Same as  ``Square'' except $\theta=-\frac{\pi}{4}$.
\item Multiplicative Noise $(\mbx,\mby) \in \Real^{p} \times \Real^{p}$: $U \sim \mc{N}(0, I_{p})$,
\begin{align*}
\mbx &\sim \mc{N}(0, I_{p}),\\
\mby_{[d]}&=U_{[d]}\mbx_{[d]} \mbox{  for $d=1,\ldots,p$.}
\end{align*}
\item Multimodal Independence $(\mbx,\mby) \in \Real^{p} \times \Real^{p}$: Let $U \sim \mc{N}(0,I_{p})$, $V \sim \mc{N}(0,I_{p})$, $U' \sim \mc{B}(0.5)^{p}$, $V' \sim \mc{B}(0.5)^{p}$. Then
\begin{align*}
\mbx&=U/3+2U'-1,\\
\mby&=V/3+2V'-1.
\end{align*}
\end{compactenum}


For the increasing dimension simulations in the main paper, we always set $\kappa=0$ and $n=100$, with $p$ increasing.  For types  $4,10,12,13,14,18,19,20$, $q=p$ such that $q$ increases as well; otherwise $q=1$.
The decaying vector $w$ is utilized for $p>1$ to make the high-dimensional relationships more difficult (otherwise, additional dimensions only add more signal).
For the one-dimensional simulations, we always set $p=q=1$, $\kappa=1$ and $n=100$.
\end{document}